\documentclass[lettersize,journal]{IEEEtran}
\usepackage{amsmath,amsfonts}
\usepackage{algorithmic}
\usepackage{algorithm}
\usepackage{array}
\usepackage[caption=false,font=small,labelfont=rm,textfont=rm]{subfig}
\usepackage{textcomp}
\usepackage{stfloats}
\usepackage{url}
\usepackage{verbatim}
\usepackage{graphicx}
\usepackage{comment}
\usepackage{multirow}
\usepackage{tikz}
\usepackage{amssymb}
\usepackage{pifont}
\usepackage{booktabs}
\usepackage[colorlinks=true,allcolors=blue]{hyperref}
\usepackage{balance}

\begin{document}
	
	\title{DRL: Discriminative Representation Learning with Parallel Adapters for Class Incremental Learning}

	\author{
		Jiawei Zhan$^{\dagger}$,
		Jun Liu$^{\dagger}$, 
		Jinlong Peng,
		Xiaochen Chen,
		Bin-Bin Gao,
		Yong Liu,\\
		and Chengjie Wang$^{\ast}$,~\IEEEmembership{Member,~IEEE}
		
		\thanks{
			Manuscript received Oct 9, 2025; revised Oct 9, 2025.
			
			J. Zhan, J. Liu, J. Peng, X.C. Chen, B.B. Gao, Y. Liu, and C.J. Wang are with Tencent Youtu Lab, Shenzhen, China.
			\emph{($^{\dagger}$Equal contributors: Jiawei Zhan and Jun Liu, $^{\ast}$Corresponding author: Chengjie Wang.)}
			
			Color versions of one or more figures in this article are available at https://doi.org/10.1109/TCSVT.2025.xxxxxxx.
			
			Digital Object Identifier 10.1109/TCSVT.2025.xxxxxxx
		}
	}
	
		\markboth{Journal of \LaTeX\ Class Files,~Vol.~XX, No.~XX, Oct~2025}%
		{Shell \MakeLowercase{\textit{et al.}}: A Sample Article Using IEEEtran.cls for IEEE Journals}
		\IEEEoverridecommandlockouts
 \IEEEpubid{\begin{tabular}[t]{@{}c@{}}0000--0000/00\$00.00~\copyright~2025 IEEE
 		Personal use is permitted, but republication/redistribution requires IEEE permission\\See https://www.ieee.org/publications/rights/index.html for more information.\end{tabular}}
		
		\maketitle
		
		\begin{abstract}
			With the excellent representation capabilities of Pre-Trained Models (PTMs), remarkable progress has been made in non-rehearsal Class-Incremental Learning (CIL) research. However, it remains an extremely challenging task due to three conundrums: increasingly large model complexity, non-smooth representation shift during incremental learning and inconsistency between stage-wise sub-problem optimization and global inference. In this work, we propose the Discriminative Representation Learning (DRL) framework to specifically address these challenges. To conduct incremental learning effectively and yet efficiently, the DRL's network, called Incremental Parallel Adapter (IPA) network, is built upon a PTM and increasingly augments the model by learning a lightweight adapter with a small amount of parameter learning overhead in each incremental stage. 
			The adapter is responsible for adapting the model to new classes, it can inherit and propagate the representation capability from the current model through parallel connection between them by a transfer gate. As a result, this design guarantees a smooth representation shift between different incremental stages. Furthermore, to alleviate inconsistency and enable comparable feature representations across incremental stages, we design the Decoupled Anchor Supervision (DAS). It decouples constraints of positive and negative samples by respectively comparing them with the virtual anchor. This decoupling promotes discriminative representation learning and aligns the feature spaces learned at different stages, thereby narrowing the gap between stage-wise local optimization over a subset of data and global inference across all classes. 
			Extensive experiments on six benchmarks reveal that our DRL consistently outperforms other state-of-the-art methods throughout the entire CIL period while maintaining high efficiency in both training and inference phases. 
		\end{abstract}
		
		
		\section{Introduction}
		\label{sec:intro}
		
		\IEEEPARstart{D}{eep} neural networks (DNNs) have demonstrated transformative success across numerous domains~\cite{he2015deepresiduallearningimage,ren2016fasterrcnnrealtimeobject,ran2022surface,Zhan_2022,li2022towards}, typically trained via supervised or self-supervised methods~\cite{he2021maskedautoencodersscalablevision} on static datasets like ImageNet~\cite{deng2009imagenet}. However, these approaches struggle to adapt effectively to data stream scenarios~\cite{ning2021rf}, necessitating incremental learning techniques such as Class-Incremental Learning (CIL)~\cite{tian2023instance,zhao2023rethinking} and incremental object detection~\cite{zhang2024learning}. Among these, {non-rehearsal CIL}~\cite{lwf,icarl,zhu2021prototype} is particularly important for privacy-sensitive applications~\cite{dong2022federated,shokri2015privacy,chamikara2018efficient} due to its strict prohibition against revisiting historical data. This setting intensifies the inherent \textit{stability-plasticity dilemma}~\cite{grossberg2012studies}, inevitably leading to catastrophic forgetting~\cite{french1999catastrophic} when acquiring new knowledge without compromising existing representations.

		\begin{figure}[t]
			\centering
			\includegraphics[width=\columnwidth]{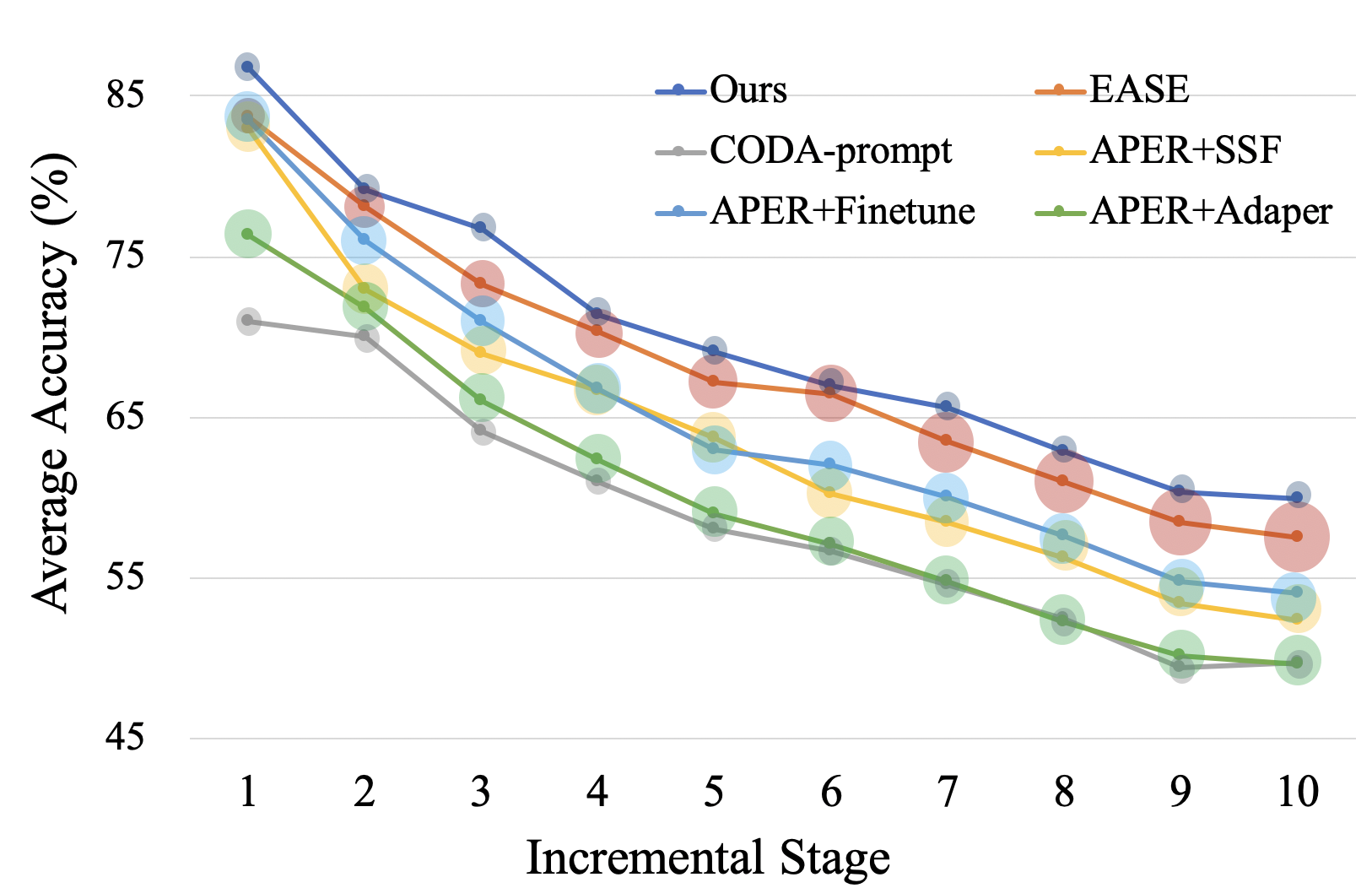}
			
			\caption{Performance comparison of different class-incremental learning methods on ImageNet-A with B0 Inc20 setting, evaluated in terms of classification accuracy versus inference complexity measured by model size. The circle sizes represent the model size during inference phase, demonstrating the efficiency advantages of our DRL framework.}
			
			\label{fig:efficiency--}
		\end{figure}
		To address this challenge, several network-based methods have been proposed, including combining multiple networks~\cite{aljundi2017expert} and dynamically expanding sub-networks~\cite{schwarz2018progress,douillard2022dytox}. Among these, DER~\cite{yan2021dynamically} and Progressive Networks (PN)~\cite{rusu2016progressive} preserve previously trained models to alleviate catastrophic forgetting while expanding newly learnable models at each incremental stage to enhance {plasticity}. FOSTER~\cite{wang2022foster} addresses the increasing model complexity in DER and PN by employing knowledge distillation to compress the model and limit its size. However, this approach requires additional training steps, increasing computational overhead.
		\IEEEpubidadjcol
		
\begin{figure*}[t]
			\centering
			\includegraphics[width=\linewidth]{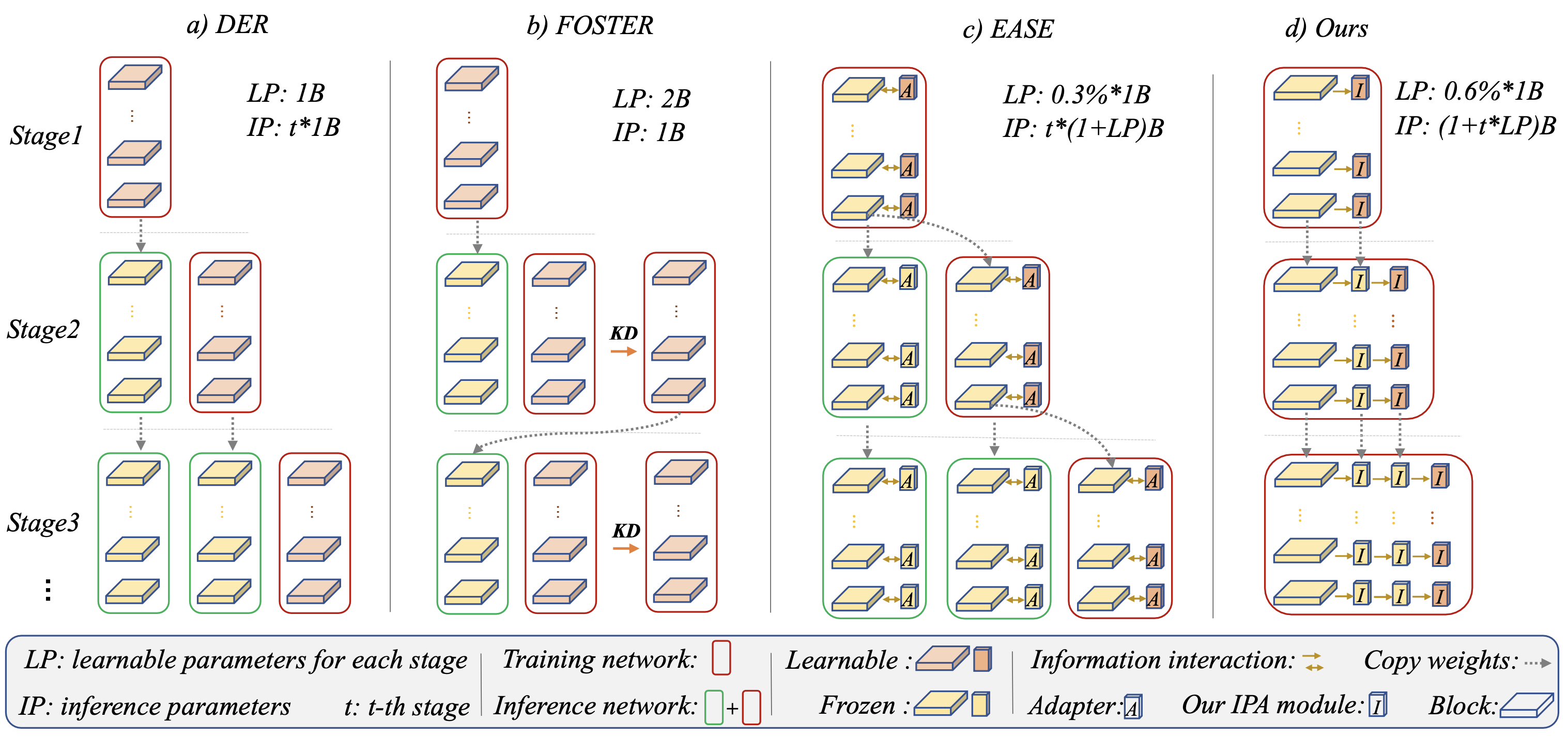}
			\caption{\textbf{Comparison of different CIL methods.} We select three representative methods: DER, FOSTER, and EASE. `1B' denotes the total parameter of a model (e.g., ViT-B/16). a) DER and b) FOSTER face the increasingly large model complexity during the training. c) EASE faces  an increasing computing overhead during inference, as well as non-smooth representation shift due to the independent training at different incremental stages. d) Our network mitigates these problems through the IPA module, which consists of a lightweight adapter with minimal parameter learning overhead at each incremental stage and a transfer gate to ensure a smooth representation shift.}
		\label{fig:overview}
	\end{figure*}

		Recent studies~\cite{zhou2024continual,zheng2023learn,li2023cross} have demonstrated that leveraging large Pre-Trained Models (PTMs) can significantly enhance CIL performance. Building on this insight, Zhou et al. proposed EASE~\cite{ease}, which achieves state-of-the-art (SOTA) performance by expanding an independent PTM with a learnable adapter~\cite{lora} to acquire new knowledge. However, EASE faces increasing computational overhead during inference (see Fig.~\ref{fig:overview}c) as it retains all previously trained models in memory to extract corresponding features. Furthermore, because DER, FOSTER, and EASE maintain independent models across different incremental stages, there is insufficient interaction between models from different stages (see Fig.~\ref{fig:overview}a-c). This isolation prevents the current model from inheriting features from previously trained models, resulting in non-smooth representation shifts. Additionally, most CIL methods rely solely on Cross-Entropy Loss for supervision at each incremental stage, which may lead to inconsistency between stage-wise local optimization on subset data and global inference across all classes that requires more discriminative representations.
		
		To overcome these limitations, we propose a Discriminative Representation Learning (DRL) framework that achieves superior performance with high efficiency (see Fig.~\ref{fig:efficiency--}). The DRL framework comprises two key components: an Incremental Parallel Adapter (IPA) network and a Decoupled Anchor Supervision (DAS) strategy. The IPA network builds upon a PTM and progressively enhances the model by learning a lightweight adapter at each incremental stage. This adapter inherits and propagates representation capability from the current model through parallel connections regulated by a learnable transfer gate, enabling smooth representation transitions with exceptional efficiency. To address the optimization inconsistency issue, DAS decouples constraints for positive and negative samples by comparing them with a virtual anchor: it enforces positive logits to exceed a fixed anchor $k$ while constraining negative logits to remain below k. This decoupling promotes discriminative representation learning and aligns feature spaces across different stages, thereby narrowing the gap between stage-wise local optimization and global inference across all classes.
		
		Extensive experiments on six benchmark datasets validate the SOTA performance of our DRL framework. On ImageNet-A, our method achieves an accuracy of 68.96\%, surpassing the current SOTA by 3.62\%. On VTAB and ObjectNet, we achieve accuracies of 95.73\% and 72.69\%, representing improvements of 2.12\% and 1.85\% over existing SOTA methods, respectively. The main contributions of this work are:
		\begin{itemize}
			\item We propose the novel IPA network, which achieves high training-inference efficiency and enables smooth representation transitions through a lightweight adapter and learnable transfer gate.
			\item We introduce DAS, a simple yet effective supervision strategy that mitigates optimization inconsistency by decoupling constraints for positive and negative samples using a fixed anchor, effectively aligning feature spaces across incremental stages. This component can be seamlessly integrated into other CIL methods.
			\item Our approach sets a new state-of-the-art across a comprehensive suite of six benchmarks, encompassing both standard CIL evaluations and challenging out-of-distribution benchmarks characterized by substantial domain shifts from the PTM's original training data.
		\end{itemize}
		
		The remainder of this paper is organized as follows. Section~\ref{s-2} reviews related work on CIL methods. Section~\ref{s-3.1} presents the preliminaries, while Sections~\ref{s-3.2} and~\ref{s-3.3} detail our proposed IPA network and DAS strategy, respectively. Comprehensive experimental evaluations are provided in Section~\ref{s-4}. Finally, we discuss the limitations of our approach, suggest future research directions, and conclude with a summary of our contributions.

	\section{Related Work}
	\label{s-2}
	\noindent\textbf{Class Incremental Learning (CIL).} CIL is essential for data streaming applications, enabling models to learn from sequentially arriving data. Methods are typically categorized based on the accessibility of training data from previous stages: \textit{rehearsal-based CIL}, which retains a small set of old exemplars in memory to mitigate catastrophic forgetting~\cite{aljundi2019gradient,liu2020mnemonics,zhao2021memory,chaudhry2018efficient,9939005}, and \textit{non-rehearsal CIL}, which fine-tunes models without relying on exemplars~\cite{simon2021learning,ewc,aljundi2018memory,zhao2020maintaining,sdc,shi2022mimicking,ceat2025,10579846,10480689}. While rehearsal-based methods demonstrate effectiveness in preserving knowledge, storing exemplars raises significant concerns regarding data security and privacy~\cite{shokri2015privacy,dong2022federated}. Consequently, research has increasingly focused on non-rehearsal CIL, which presents a more challenging but privacy-preserving alternative.

	\noindent\textbf{Network-based Methods.} As a prominent category within non-rehearsal CIL~\cite{qu2021recent}, these methods allocate specific model parameters to each stage to address catastrophic forgetting. Recent {expandable network} approaches~\cite{wang2022foster,douillard2022dytox,chen2023dynamic} have demonstrated particularly strong performance. However, many methods in this category (e.g., DER~\cite{yan2021dynamically} and Progressive Networks~\cite{rusu2016progressive}) suffer from expanding the backbone network or incorporating large modules~\cite{10568189}, leading to increasingly cumbersome networks that lack flexibility and efficiency after multiple incremental stages. This architectural inflation poses practical limitations for long-term deployment scenarios.

	\noindent\textbf{Pre-Trained Model-based Methods.} PTM-based methods leverage the strong representational capabilities of pre-trained models and have become a prominent research direction recently~\cite{zhou2024continual,wang2024hierarchical,mcdonnell2024premonition}. These approaches are generally divided into prompt-based~\cite{coda,wang2022s} and adapter-based methodologies. Recent methods such as L2P~\cite{l2p} and DualPrompt~\cite{wang2022dualprompt} utilize prompt tuning based on PTMs for incremental learning tasks. However, these methods typically maintain a unified prompt pool that requires continuous updating, which can lead to prompt-level forgetting and restricted representation ability over time. Alternative approaches, including APER~\cite{aper} and EASE~\cite{ease}, expand an independent PTM with trainable adapters~\cite{adapter,lora} to fine-tune the model for each stage. While effective, these methods often utilize all stage models during inference, resulting in computational inefficiency and potential representation shifts due to limited interaction between models across different stages.

	\noindent\textbf{Supervision in Class Incremental Learning.} Existing continual learning methods employ diverse supervisory signals, spanning classification losses (e.g., cross-entropy~\cite{ease} and margin-based formulations like CosFace~\cite{wang2018cosface}) to regularization mechanisms (e.g., knowledge distillation~\cite{fcs,lwf,hinton2015distilling}). Notably, softmax cross-entropy remains the predominant choice in stage-wise training paradigms across leading methods including EASE, RanPAC~\cite{mcdonnell2023ranpac}, and SAFE~\cite{zhao2024safe}. Nevertheless, these approaches exhibit an intrinsic limitation: inconsistent separation granularity between training objectives and inference requirements. The local optimization at each stage focuses only on the current class subset, while global inference must discriminate across all encountered classes. To resolve this misalignment, we propose {Decoupled Anchor Supervision (DAS)}, which introduces a fixed anchor to decouple constraints on positive and negative samples. This framework simultaneously eliminates the non-independence inherent in conventional losses while ensuring consistent decision boundaries during deployment, ultimately yielding significant performance improvements across incremental learning scenarios.

	\section{Method}
	\label{s-3}
	
	\subsection{Preliminaries} 
	\label{s-3.1}
	CIL aims to train a model with training samples arriving in sequence. 
	This incremental process can be divided into $\mathit{T}$ stages. For stage $\mathit{t}$ $\in$ \{1, 2, $\cdots$, $\mathit{T}$\}, the training samples belonging to the stage $\mathit{t}$ are represented as $\mathit{D^{t}} = \{\mathit{X^{t}},\mathit{Y^{t}} \}$, where $\mathit{X^{t}}$ is the input data, and $\mathit{Y^{t}}$ corresponds to the associated label. 
	The classes across different stages do not overlap, i.e., $\mathit{Y^{1}} \cap \mathit{Y^2} \cap \cdots \cap \mathit{Y^{T}} = \emptyset$. 
	The non-rehearsal CIL satisfies $\mathit{D^1} \cap \mathit{D^2} \cap \cdots \cap \mathit{D^T} = \emptyset$, meaning no previous data is stored for replay, making this the most challenging and practical CIL scenario.
	During the training of the model at the $t$-th stage, we can only access the data $D^t$, while the stage identity $\mathit{t}$ is not available during inference. After each stage, the trained model is evaluated on all previously seen classes, i.e., $Y^1 \cup Y^2 \cup \cdots \cup Y^{t}$, which requires the model to balance stability (remembering old classes) and plasticity (learning new classes) effectively.

	The model of CIL can be formulated as $f(\mathbf{x}) = X \rightarrow Y$, which aims to minimize the empirical risk:
\begin{equation} 
	\label{eq:risk}
	\sum\nolimits_{(\mathbf{x}, y) \in \mathit{D^1} \cdots \cup \mathit{D^T} } L(f(\mathbf{x}),y),
\end{equation}
	Here, we decouple our model into embedding module $\Phi(\cdot):\ \mathbb{R}^D \rightarrow \mathbb{R}^d$ and classifier layer $\mathbf{W} \in \mathbb{R}^{d \times |Y|}$, where $d$ represents the embedding dimension and $Y$ represents the label space. The model output is then denoted as $f(\mathbf{x})=\mathbf{W}^{\top} \Phi(\mathbf{x})$. 
	Since our DRL is based on PTM, for the $t$-th stage, the embedding module can be further parameterized as $\Theta = \{\theta_t^o, \theta_t^n\}$, where $\theta_t^o $ and $\theta_t^n$ are the parameters of the previously trained model (e.g., PTM) and the new expanding network (e.g., adapter~\cite{ease}), respectively. 
	Furthermore, for the $l$-th  transformer block~\cite{dosovitskiy2020image}, where $l \in \{1, \cdots,L \}$ and $L$ represents the total number of blocks (e.g., $L = 12$ in ViT-B/16), the parameters are denoted as $\{ \theta_t^{o_l}, \theta_t^{n_l} \}$. 
	The classifier layer can be further decomposed into a combination of $\mathbf{W} = [\mathbf{w}_1, \cdots,\mathbf{w}_{|Y|}]$. The classifier weight for class $i$ is $\mathbf{w}_i$, where $\mathbf{w}_i \in \mathbb{R}^{d \times 1}$.

	Following the EASE~\cite{ease}, during the training phase, the logit for class $i$ given by:
\begin{equation}
	\label{eq:logit}
	z_{i} =  s \cdot \cos(\mathbf{w}_i,\Phi(\mathbf{x})),
\end{equation}
	where $s$ is a learnable scale factor during the training phase. The logit $z_i$ is passed to the softmax function to obtain the output probability:
	\begin{equation}
		\label{eq:softmax}
    p_i=\frac{e^{z_i}}{\sum_{j} e^{z_j}}=\frac{e ^{s \cdot \cos(\mathbf{w}_i, \Phi(\mathbf{x})) }} {\sum_{j} e ^{s \cdot \cos(\mathbf{w}_j, \Phi(\mathbf{x}))}},
	\end{equation}
	During inference, the prototype-based classifier extracts the final \texttt{[CLS]} token in ViT-B as the class center $\mathbf{c}_i$ (i.e., prototype) for the $i$-th class and directly replaces $\mathbf{w}_i$. It then utilizes cosine distance to calculate the probability as:
	\begin{equation}
		\label{eq:cosine}
    \hat{p}_{i} = \cos(\mathbf{c}_i, \Phi(\mathbf{x})) = {\frac{\mathbf{c}_i^{\top} \Phi(\mathbf{x})}{\| \mathbf{c}_i \|_{2} \cdot \|\Phi(\mathbf{x})\|_{2}}}.
	\end{equation}
	This prototype-based inference strategy enhances robustness by leveraging class centroids that are less susceptible to the representation drift that can occur during incremental learning.

	\begin{figure}[t]
		\centering
		
		\includegraphics[width=\columnwidth]{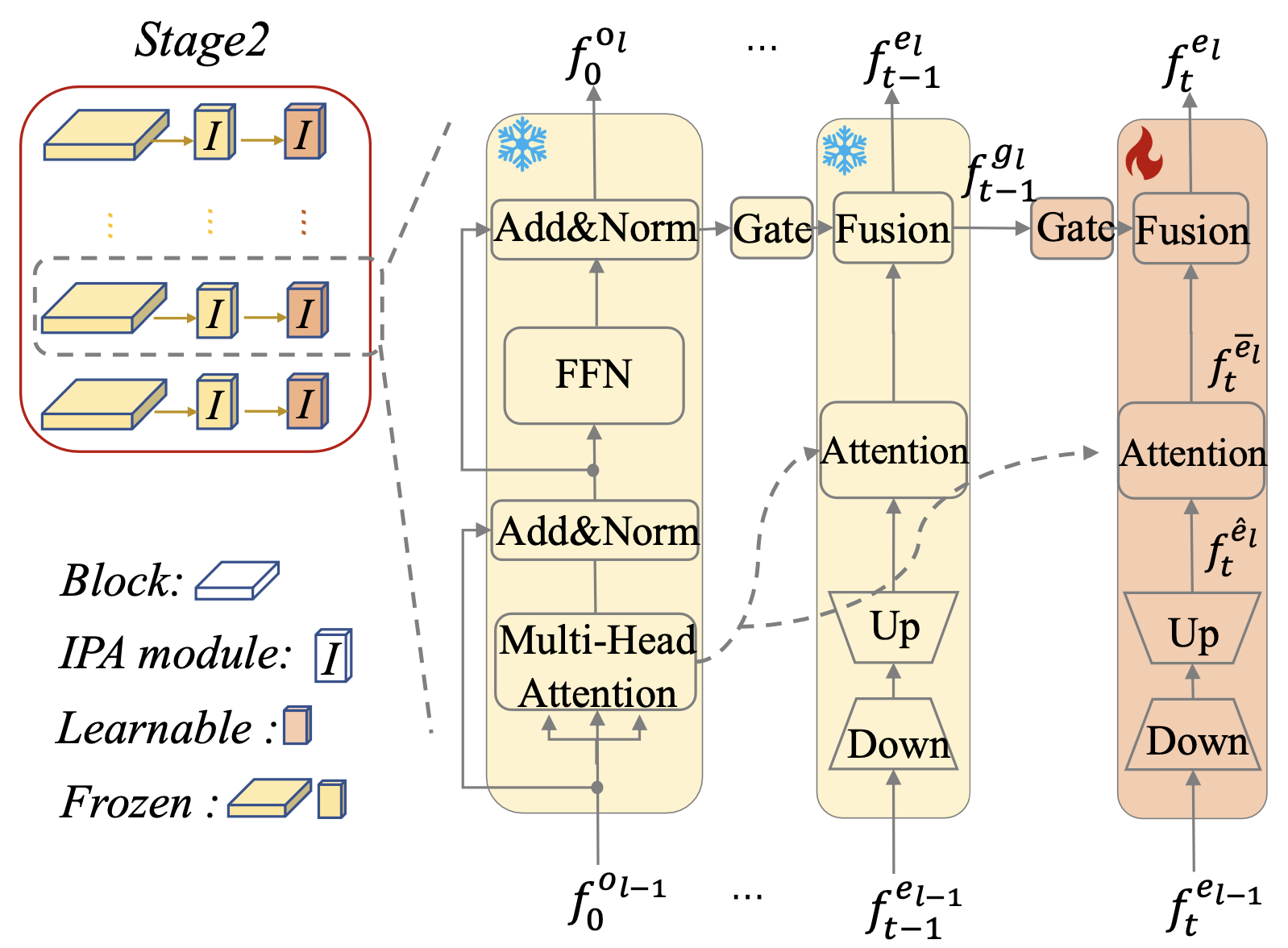}
		
		\caption{\textbf{Detailed architecture of the Incremental Parallel Adapter (IPA) network.} This design incorporates a lightweight adapter structure, a reusable attention mechanism, and a transfer gate to enable seamless knowledge propagation across incremental learning stages.}

		\label{fig:ppl2}
	\end{figure}

	\subsection{Incremental Parallel Adapter (IPA)}
	\label{s-3.2}
	To address the large model complexity during training or inference phase~\cite{yan2021dynamically, wang2022foster,ease} and non-smooth representation shift caused by the independent training of models at different stages, we propose Incremental Parallel Adapter (IPA) network. Inspired by PTM-based methods that demonstrate promising performance in CIL, our network is built upon a PTM and increasingly expands a IPA module in each block for current incremental stage, leading to  superior performance with high efficiency. 
	The core innovation lies in maintaining a fixed backbone while progressively integrating trainable parallel adapters, ensuring parameter efficiency while enabling continuous adaptation.

	\noindent\textbf{Overview of the IPA Network.}
	The IPA network is illustrated in Fig.~\ref{fig:overview}d, the details of each block are shown in Fig.~\ref{fig:ppl2}, which mainly consists of two parts: the previously trained model parameterized by $\theta^o = \{\theta^{o_l}\}_{l=1}^L$ and the IPA module parameterized by $\theta^n = \{\theta^e,\theta^g\}$. The previously trained model $\theta^o$ is utilized to extract fundamental features, which can be either a PTM (in the first stage) or a model trained in the previous stage (the stage after the first one). This frozen shared backbone preserves established representations and prevents catastrophic forgetting.
	On the other hand, the {IPA module} consists of an efficient lightweight adapter parameterized by $\theta^e = \{\theta^{e_l}\}_{l=1}^L$ and a learnable transfer gate parameterized by $\theta^g = \{\theta^{g_l}\}_{l=1}^L$ . 
	For the $l$-th block during training, $\theta^{o_l}$ is fixed to retain representation ability and mitigate catastrophic forgetting while $\theta^{n_l}$ (comprising $\theta^{e_l}$ and $\theta^{g_l}$) is learnable to acquire new knowledge. This design ensures that the model maintains stability through the fixed backbone while achieving plasticity via the adaptable IPA modules.

	\noindent\textbf{Lightweight Adapter in the IPA Module.}
	The lightweight adapter consists of two 1x1 convolutional layers followed by a reusable attention module. Specifically,  the input of the $l$-th adapter for the $t$-th stage is the output of the  $(l-1)$-th block (e.g., $\boldsymbol{f}_t^{e_{l-1}}$ in Fig.~\ref{fig:ppl2}).
	$\boldsymbol{f}_t^{e_{l-1}}$ is passed to a 1x1 convolutional layer $\mathbf{W}_{down} \in \mathbb{R}^{d \times r}$ for downsampling, followed by an activation function, and then another 1x1 convolutional layer $\mathbf{W}_{up} \in \mathbb{R}^{r \times d}$ for upsampling. The output is denoted as $\boldsymbol{f}_t^{\hat{e}_l}$.  This bottleneck-like structure, consisting solely of two 1x1 convolutions, makes it extremely lightweight, adding minimal parameters while enabling feature transformation.
	The reusable attention module aims to enhance the correlations between features (or tokens) without introducing additional learnable parameters. Traditionally, the attention module~\cite{vaswani2023attentionneed} requires three additional 1x1 convolutional layers to generate the $\mathbf{Q},\ \mathbf{K},\ \mathbf{V}$, and utilizes $\mathbf{Q}$ and $\mathbf{K}$ to calculate the attention matrix $\mathbf{A}^e$, i.e., $\mathbf{A}^e = softmax(\frac{\mathbf{Q}\mathbf{K}^{\top}}{\sqrt{d_1}})$, where $d_1$ is the dimension of $\mathbf{Q}$ and $\mathbf{K}$. 
	However, this traditional approach results in additional learnable parameters. Given that the PTM inherently possesses strong representational capabilities, and the attention matrix $\mathbf{A}^o$ in the PTM encapsulates the relationships among features.
	We propose that the $\mathbf{A}^e$ can be replaced by the one in the PTM (i.e., $\mathbf{A}^e = \mathbf{A}^o$) without losing its {plasticity}. This reuse of pre-computed attention patterns significantly reduces computational overhead while maintaining expressive power.
	Finally, $\boldsymbol{f}_t^{\hat{e}_l}$ is treated as $\mathbf{V}$, and the output is computed as $\boldsymbol{f}_t^{\bar{e}_l} = \mathbf{A}_t^{o_l} \boldsymbol{f}_t^{\hat{e}_l}$. 
	This attention module, which we call ``reusable attention", serves as a unique form of cross-attention between the PTM and the new IPA module, effectively transferring structural knowledge from the pre-trained model to the adapter.

	\noindent\textbf{Transfer Gate in the IPA Module.}
	The learnable transfer gate is responsible for connecting two parallel adapters between two adjacent stages to ensure a smooth representation shift.
	A naive approach would be to directly sum the old and new features. However, we have found that shallow and deep layers in the previously trained model exhibit different characteristics, and the adapter should selectively inherit knowledge from these layers. Therefore, we design a learnable transfer gate for each block to preserve essential knowledge and enhance {plasticity}.  
	Specifically, the gate includes downsample and upsample layers identical to those in the lightweight adapter, followed by a sigmoid function that constrains its output to a range between 0 and 1. 
	The input of the $l$-th gate for task $t$ is denoted as $\boldsymbol{f}_{t-1}^{g_l} = (1-\gamma)\boldsymbol{f}_{t-1}^{e_l} + \gamma \boldsymbol{f}_{0}^{o_l}$. Here, the $\boldsymbol{f}_{0}^{o_l}$ represents the feature of $l$-th block in the PTM, and the output is a weight mask $\mathbf{M}_t^{l}$. 
	Finally, we fuse the features of the previously trained network and the adapter as follows: $\boldsymbol{f}_t^{e_l} = (1-\mathbf{M}_t^l) \boldsymbol{f}_t^{\bar{e}_l}  + \mathbf{M}_t^l \boldsymbol{f}_{t-1}^{g_l}$. This adaptive fusion allows the model to dynamically balance contributions from current features versus historical knowledge at a per-block granularity.

	Generally, the IPA module can be integrated into all blocks. However, we have found that fusing the features of the last block reduces {plasticity}. 
	Therefore, we use a transfer block to replace the IPA module of the $L$-th layer and independently introduce two lightweight linear layers instead of the original Feedforward Network (FFN) in this block. This specialized handling of the final block preserves the model's capacity to learn new representations while maintaining efficiency.

	During inference, performing one inference on the $t$-th stage model is sufficient to obtain the embedding representation $ \mathbf{F}_t = [\boldsymbol{f}_0^{o_L},  \boldsymbol{f}_1^{e_L}, \cdots ,\boldsymbol{f}_t^{e_L}]$. 
	Following EASE~\cite{ease}, we employ the ``semantic-guided prototype" to synthesize new features for old classes without accessing any old class instance and classify them using Formula~\ref{eq:cosine}.

	\subsection{Decoupled Anchor Supervision (DAS)}
	\label{s-3.3}
	\noindent\textbf{Inconsistency between stage-wise local optimization and global inference.} 
	A typical approach for training our proposed IPA model is to perform stage-wise gradient descent using the standard cross-entropy loss, where model optimization is performed relative to the class activations at the current stage.
	This introduces a key issue: as the logit scale drifts across training stages, the learned decision boundary becomes stage-dependent. Specifically:
	\begin{itemize}
		\item During each stage's optimization, feature activations and logits are calibrated {relative to the mean activation intensity} $\mu_t = \mathbb{E}_{i \in \mathcal{C}_t}[\|\mathbf{w}_i\|]$ of that stage's classes $\mathcal{C}_t$, creating stage-specific reference points. 
		\item During global inference, features from different stages must be compared directly despite lacking a consistent reference point. This incompatibility results in misaligned and suboptimal decision boundaries
	\end{itemize}
	This reference point inconsistency prevents meaningful comparison of logits across stages, undermining the model's ability to maintain discriminability across all learned classes.
	To alleviate this problem, we propose the {Decoupled Anchor Supervision (DAS)} to optimize the representation learning and enhance class cross-stage compatibility. 
	
	Specifically, DAS introduces an {predefined virtual anchor} $k$ that serves as a fixed reference point across all incremental stages and decouples constraints for positive and negative samples, enabling consistent logit interpretation and feature space alignment. 
	The core objective of DAS is to enforce the following anchor-based constraints:
	\begin{align}
    \text{Positive:}\ & z_i > k, \\
\text{Negative:}\ & z_j < k\quad \text{for}\ j \neq i.
\label{eq:constraints}
	\end{align}
	where $z_i = s \cdot \cos(\mathbf{w}_i, \Phi(\mathbf{x}))$ is the logit for class $i$. 
	
	To achieve these constraints, we design the following decoupled supervision signals:
	\begin{equation}
		\label{eq:ga}
    L^{pos}= -\sum\nolimits_{i} y_{i} \log({p}_i^{pos}),\ \
L^{neg}= -\log(p^{neg}),
	\end{equation}
	Here, $y_i$ is the groundtruth label, and the $p_i^{pos}$ and $ p^{neg}$ calculate as follows:
	\begin{equation}
		\label{eq:score_ga_pos}
    p_i^{pos}=\frac{e^{z_i}}{ e^{z_i} + e^{k}} = \frac{e ^{s \cdot \cos(\mathbf{w}_i, \Phi(\mathbf{x})) }} {e^{s \cdot \cos(\mathbf{w}_i,\Phi(\mathbf{x}))} + e^{k}},
	\end{equation}
	\begin{equation}
		\label{eq:score_ga_neg}
    p^{neg} = \frac{e^{k}}{ \sum_{j,j\neq i}^C e^{z_j} + e^{k}} = \frac{e^{k}} {\sum_{j,j \neq i} e^{s \cdot \cos(\mathbf{w}_j, \Phi(\mathbf{x}))} + e^{k}}.
	\end{equation}
	Additionally, most CIL tasks involve single-label classification. Therefore, we can simplify Equation~\ref{eq:ga} for single-label classification tasks as follows:
	\begin{equation}
		\label{eq:ga_final}
    L^{pos} = -\log(p^{pos}),\ \ L^{neg} = -\log(p^{neg}),
	\end{equation}
	To balance $L^{pos}$ and $L^{neg}$, we set  $\lambda_p, \lambda_n$ as the loss weight, and our proposed DAS is defined as:
	\begin{equation}
		\label{eq.7}
    L_{das} = \lambda_p L^{pos} + \lambda_n L^{neg}.
	\end{equation}
	
	\begin{figure}[t]
		\centering
		\includegraphics[width=0.95\columnwidth]{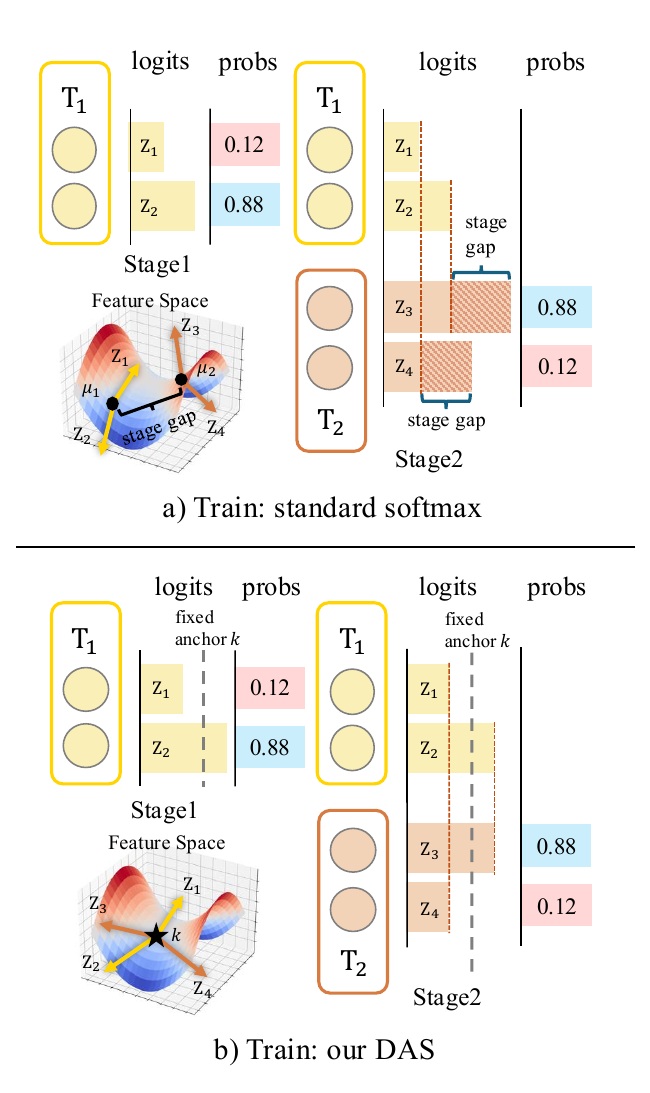}
		\caption{\textbf{Comparisons of decision boundaries.} (a) Standard softmax exhibits inconsistent separation boundaries across incremental stages; (b) Our proposed DAS establishes consistent decision boundaries through a fixed virtual anchor, enabling stable representation learning across all incremental process.}
		\label{fig:ce-ga}
	\end{figure}
	\noindent\textbf{Analysis of DAS.}
	Fig.~\ref{fig:ce-ga} illustrates a fundamental limitation of standard softmax-based training in incremental learning and its resolution via Decoupled Anchor Supervision (DAS). In standard softmax training (Fig.~\ref{fig:ce-ga}a, upper), classifiers for sequential tasks are optimized independently. While this approach achieves local discriminability {within} each individual task (e.g., within Task 1, the predicted class is determined if its logit exceeds the logits of all other classes within Task 1), it suffers from {decision boundary drift} due to semantic shifts and feature distribution misalignment {across tasks}. For instance, the boundary separating $z_3$ and $z_4$ in Task 2 drifts relative to the boundary established for $z_1$ and $z_2$ in Task 1 (Fig.~\ref{fig:ce-ga}a, upper). This drift introduces task-specific biases into the feature/logit spaces, undermining global discriminability and distorting distance-based metrics (e.g., prototype matching).
	{DAS (Fig.~\ref{fig:ce-ga}b, lower)} addresses this by anchoring supervision to a {fixed reference point} $k$. For a sample $\mathbf{x}_1$ from class 1 with feature $\mathbf{f}_{x_1} = \Phi(\mathbf{x}_1)$, DAS decouples classification by requiring that the positive logit $\mathbf{w}_1^\top\mathbf{f}_{x_1}$ exceeds the anchor $k$, while simultaneously enforcing that all negative logits $\mathbf{w}_j^\top\mathbf{f}_{x_1}$ ($j \neq 1$) fall below this same anchor value. Crucially, $k$ is task-agnostic and fixed (e.g., $k=0$), establishing an {absolute reference axis} that disentangles supervision from task-specific contexts. This unified anchoring ensures all task-specific classifiers ($\mathbf{w}_i$) and features are implicitly aligned to the same global reference, guaranteeing {cross-task compatibility} in both logit and feature spaces (Fig.~\ref{fig:ce-ga}b), thus stabilizing representation learning (Table~\ref{tab:ablation-loss-others}).  
	Besides, DAS naturally accommodates {discriminative margins}: replacing $k$ with \textit{separate anchors} $k_+$ and $k_-$ introduces a margin $\delta=k_+$-$k_-$\textgreater0. This can create a ``buffer zone'' that further {enhances feature separability}.
	
	\begin{table*}[t]
		\caption{
		\textbf{Comparison of average and final accuracy across six benchmark datasets.} All methods, implemented without exemplars and using the {ViT-B/16-IN21K} pretrained weights, are fairly compared on ImageNet-R/A (IN-R/A) and other benchmarks. The best performances are highlighted in \textbf{bold}.
		}
		
		\label{tab:benchmark}
		\centering
		\resizebox{1.0\textwidth}{!}{%
			\begin{tabular}{lccccccccccccccccc}
				\toprule
				\multicolumn{1}{l}{\multirow{2}{*}{Methods}} & 
				\multicolumn{2}{c}{CIFAR B0 Inc5} 
				& \multicolumn{2}{c}{IN-R B0 Inc5}
				& \multicolumn{2}{c}{IN-A B0 Inc20}
				& \multicolumn{2}{c}{ObjNet B0 Inc10}
				& \multicolumn{2}{c}{Omni B0 Inc30}
				& \multicolumn{2}{c}{VTAB B0 Inc10} \\
				\cmidrule(lr){2-3} \cmidrule(lr){4-5} \cmidrule(lr){6-7} \cmidrule(lr){8-9} \cmidrule(lr){10-11} \cmidrule(lr){12-13}
				& {$\bar{\mathcal{A}}$} & ${\mathcal{A}_T}$
				& {$\bar{\mathcal{A}}$} & ${\mathcal{A}_T}$   
				& {$\bar{\mathcal{A}}$} & ${\mathcal{A}_T}$ 
				& {$\bar{\mathcal{A}}$} & ${\mathcal{A}_T}$ 
				& {$\bar{\mathcal{A}}$} & ${\mathcal{A}_T}$ 
				& {$\bar{\mathcal{A}}$} & ${\mathcal{A}_T}$ 
				\\
				\midrule
				Finetune	                         & 38.90 & 20.17 &21.61 & 10.79 &24.28 & 14.51 & 19.14 & 8.73 & 23.61 & 10.57 & 34.95 & 21.25  \\
				Finetune Adapter~\cite{adapt_former} & 60.51 &49.32 & 47.59 &40.28 &45.41 &41.10 &50.22 &35.95 &62.32& 50.53 &48.91 & 45.12 \\
				LwF~\cite{lwf}                     & 46.29 & 41.07 & 39.93 &26.47 &37.75 & 26.84 & 33.01 & 20.65 & 47.14 &33.95 & 40.48 & 27.54\\
				SDC~\cite{sdc}               & 68.21 &63.05 & 52.17 & 49.20 & 29.11 & 26.63 & 39.04 & 29.06 &60.94 & 50.28 &45.06 &22.50\\
				L2P~\cite{l2p}                      & 85.94 & 79.93 & 66.53 & 59.22 & 49.39 & 41.71 & 63.78 & 52.19 &73.36 & 64.69 & 77.11 & 77.10\\
				DualPrompt~\cite{wang2022dualprompt} & 87.87 & 81.15 & 63.31 & 55.22 & 53.71 & 41.67 & 59.27 & 49.33 & 73.92 & 65.52 & 83.36 & 81.23\\
				CODA-Prompt~\cite{coda}             & 89.11 & 81.96 & 64.42 &55.08 & 53.54 & 42.73 & 66.07 &53.29 &77.03 &68.09 &83.90 &83.02\\
				SimpleCIL~\cite{aper}             & 87.57 & 81.26  & 62.58 & 54.55 & 59.77 & 48.91 & 65.45 & 53.59 & 79.34 & 73.15 & 85.99 & 84.38\\
				APER w/ Adapter~\cite{aper}  & 90.65 &  85.15  & 72.35 & 64.33 & 60.47 &49.37 & 67.18 & 55.24 & 80.75 & 74.37 & 85.95 & 84.35\\
				EASE~\cite{ease}                    & 91.51 &  85.80  & 78.31& 70.58 & 65.34 & 55.04 & 70.84 & 57.86 & 81.11& 74.85& 93.61& 93.55\\
				\midrule
				DRL        & \bf 92.01 & \bf 86.91 & \bf 78.87 & \bf 72.20 & \bf 68.96 & \bf 59.38 &\bf 72.69 & \bf 60.29 &\bf 81.26 &\bf 74.98 & \bf 95.73 &\bf 95.01 \\
				\bottomrule
			\end{tabular}
			
		}
	\end{table*}

	\begin{table}[t]
		\caption{
			\textbf{Comparison of average and final accuracy across two benchmark datasets.} All methods, implemented without exemplars and using the ViT-B/16-IN1K pretrained weights, are fairly compared on ImageNet-R/A (IN-R/A) and other benchmarks. The best performances are highlighted in \textbf{bold}.
		}
		
		\label{tab:benchmark-typicalmethods}
		\centering
		\resizebox{1.0\columnwidth}{!}{%
			\begin{tabular}{lcccccc}
				\toprule
				\multicolumn{1}{l}{\multirow{2}{*}{Methods}} & 
				\multicolumn{2}{c}{ObjNet B0 Inc20} & \multicolumn{2}{c}{ImageNet-A B0 Inc20}  \\
				\cmidrule(lr){2-3} \cmidrule(lr){4-5}
				& {$\bar{\mathcal{A}}$} & ${\mathcal{A}_T}$  
				& {$\bar{\mathcal{A}}$} & ${\mathcal{A}_T}$	\\
				\midrule
				iCaRL~\cite{icarl}          & 33.43 &19.18 &29.22 & 16.16\\
				LUCIR~\cite{hou2019learning}  & 41.17 &25.89 &31.09 &18.59\\
				DER~\cite{yan2021dynamically} & 35.47 &23.19 &33.85 & 22.27 \\
				FOSTER~\cite{wang2022foster} & 37.83 &25.07 &34.82 & 23.01\\
				MEMO~\cite{memo}                  & 38.52 &25.41 &36.37 &24.46\\
				FACT~\cite{zhou2022forward}   & 60.59 &50.96 &60.13 & 49.82\\
				SimpleCIL~\cite{aper}         & 62.11 &51.13 &59.67 & 49.44\\
				APER w/ SSF~\cite{aper}       & 68.75 &56.79 &63.59 & 52.67 \\
				EASE~\cite{ease}            & 70.44 &58.37 &65.74 & 57.28 \\
				\midrule
				DRL                          & \bf 72.81 & \bf 61.00  & \bf 69.42   & \bf  59.97 \\
				\bottomrule
			\end{tabular}
		}
		
	\end{table}
	
	Given feature distributions inherited from pretrained models exhibit advantageous generalization properties, we incorporate knowledge distillation to constrain feature space during incremental updates. The final optimization objective integrates DAS with distillation regularization:
	\begin{equation}
    L_{total} = L_{das} + \alpha L_{kd}.
	\end{equation}
	Here, $\alpha$ is the loss weight for $L_{kd}$, $L_{kd}$ is the loss of the final embedding (such as the final \texttt{[CLS]} token in ViT) between the PTM and the newly inserted IPA module. For simplicity, we use cosine distance as the metric for $L_{kd}$.
	\section{Experiments}
	\label{s-4}
	In this section, we comprehensively evaluate DRL against state-of-the-art methods on six benchmark datasets, utilizing diverse pre-trained models and data splits. This demonstrates DRL's superior performance. Ablation studies confirm its effectiveness, and visual analyses provide additional insights.

	\noindent\textbf{Datasets.}
	We evaluate the performance on six datasets: CIFAR100~\cite{cifar}, ImageNet-R~\cite{imagenetr}, ImageNet-A~\cite{imageneta}, ObjectNet~\cite{objectnet}, OmniBench~\cite{omni}, and VTAB~\cite{vtab}. This selection encompasses both standard class-incremental learning benchmarks (CIFAR100, ImageNet-R) and challenging out-of-distribution datasets (ImageNet-A, ObjectNet, OmniBench, VTAB) that exhibit significant domain shifts from the pre-training data. The datasets vary in scale and complexity: CIFAR100 contains 100 classes, ImageNet-R and ImageNet-A each have 200 classes, ObjectNet comprises 200 classes, OmniBench includes 300 classes, and VTAB consists of 50 classes. For ablation studies and detailed analysis, we primarily focus on ImageNet-A (known for its challenging samples that challenge ImageNet pre-trained models) and VTAB (featuring diverse classes from multiple complex domains), as they effectively showcase the robustness of our approach under distribution shifts.
	Following established benchmark protocols~\cite{icarl,l2p,zhou2023deep}, we employ the `B-$m$ Inc-$n$' class split convention, where $m$ denotes the number of classes in the initial stage and $n$ represents the number of classes introduced in each incremental stage.
	
	\noindent\textbf{Evaluation Metric.}
	Consistent with standard evaluation practices in class-incremental learning~\cite{icarl}, we define $\mathcal{A}_t$ as the Top-1 accuracy after the $t$-th incremental stage. Our primary evaluation metrics are: (1) ${\mathcal{A}_T}$, representing the final performance after all incremental stages, which reflects the model's ability to retain knowledge across the entire learning trajectory; and (2) $\bar{\mathcal{A}}=\frac{1}{T}\Sigma_{t=1}^T \mathcal{A}_t$, the average performance across all stages, which comprehensively captures the stability-plasticity trade-off throughout the incremental learning process.

	\noindent\textbf{Comparison Methods.}
	We compare DRL framework against state-of-the-art methods from two categories: (1) {PTM-based CIL methods} including L2P~\cite{l2p}, DualPrompt~\cite{wang2022dualprompt}, CODA-Prompt~\cite{coda}, APER~\cite{aper}, and EASE~\cite{ease}, which leverage pre-trained vision transformers; and (2) {Conventional CIL approaches} such as LwF~\cite{lwf}, SDC~\cite{sdc}, iCaRL~\cite{icarl}, LUCIR~\cite{hou2019learning}, DER~\cite{yan2021dynamically}, FOSTER~\cite{wang2022foster}, MEMO~\cite{memo} and FACT~\cite{zhou2022forward}, which represent established techniques in incremental learning. All methods are initialized with identical pre-trained models to ensure fair comparison.

	\begin{table}[t]
		\caption{\textbf{Ablation study on the components of Decoupled Anchor Supervision (DAS).} The evaluation, performed on ImageNet-A and VTAB using ViT-B/16-IN21K weights, compares different loss configurations including cross-entropy ($L_{ce}$), knowledge distillation ($L_{kd}$), and DAS ($L_{das}$), demonstrating the complementary benefits of each component and the optimal combination in DRL.}
		
		\label{tab:ablation-loss}
		\centering
		\resizebox{1.0\columnwidth}{!}{%
			\begin{tabular}{lccccccc}
				\toprule
				\multicolumn{1}{l}{\multirow{2}{*}{Methods}} &  \multicolumn{3}{c}{Components} & \multicolumn{2}{l}{IN-A B0 Inc20}  & \multicolumn{2}{l}{VTAB B0 Inc10} \\ 
				\cmidrule(lr){2-4} \cmidrule(lr){5-6}\cmidrule(lr){7-8}
				&$L_{ce}$ &$L_{kd}$&$L_{das}$    & {$\bar{\mathcal{A}}$} & ${\mathcal{A}_T}$  & {$\bar{\mathcal{A}}$}  & ${\mathcal{A}_T}$  \\
				\midrule
				EASE~\cite{ease}     & \ding{51} & \ding{55} & \ding{55}     & 65.34          & 55.04         & 93.61       & 93.55      \\
				EASE~\cite{ease}     & \ding{51} & \ding{51} & \ding{55}     & 66.51           & 55.50         & 91.64       & 90.99      \\
				DRL    & \ding{51} & \ding{55} & \ding{55}     & 66.16	          & 56.05         & 94.48	       & 93.83      \\   
				DRL    & \ding{51} & \ding{51} & \ding{55}     & 67.24          & 57.12         & 94.72       & 94.03      \\
				DRL    & \ding{55} & \ding{51} & \ding{51}     & \bf 68.96      & \bf 59.38     & \bf 95.73   & \bf 95.01      \\ 
				\bottomrule
			\end{tabular}
		}
		
	\end{table}

	\noindent\textbf{Training Details.}
	Following established protocols in~\cite{l2p,aper}, we utilize Vision Transformer Base (ViT-B/16) architectures pre-trained on ImageNet-21K (IN21K) and ImageNet-1K (IN1K). For DRL, optimization is performed using SGD~\cite{robbins1951stochastic} with momentum (0.9), weight decay ($5\times10^{-4}$), and batch size 48, trained for 20 epochs per incremental stage. The learning rate follows a cosine annealing schedule starting from 0.01. Key hyperparameters are set as: $\alpha=0.5$ (loss balancing), $\lambda_p=3$ (positive anchor weighting), $\lambda_n=1$ (negative anchor weighting), and $\gamma=0.9$ (feature retention ratio). The virtual anchor mechanism uses $k=1$ neighbor, while the dimensionality reduction in adapter layers employs $r=48$ dimensions for both $\mathbf{W}_{\text{down}} \in \mathbb{R}^{d\times 48}$ and $\mathbf{W}_{\text{up}} \in \mathbb{R}^{48\times d}$. Within the final transformer block, the IPA module implements sequential linear transformations with $\mathbf{W}_{\text{first}} \in \mathbb{R}^{768\times768}$ and $\mathbf{W}_{\text{second}} \in \mathbb{R}^{768\times768}$. Unless otherwise specified, all experiments utilize ViT-B/16-IN21K as the pre-trained model for initialization to ensure consistent and fair comparisons across different benchmark evaluations.

	\begin{table}[t]
		\caption{\textbf{Comparative study of different loss functions integrated with our DRL framework.} The evaluation, performed on ImageNet-A and VTAB using ViT-B/16-IN21K weighs, includes Binary Cross-Entropy ($L_{bce}$), CosFace ($L_{cf}$), and Decoupled Anchor Supervision ($L_{das}$), highlighting DAS's superior performance in enhancing feature discriminability and cross-stage compatibility.}
		
		\label{tab:ablation-loss-others}
		\centering
		\resizebox{1.0\columnwidth}{!}{%
			\begin{tabular}{ccccccc}
				\toprule
				\multicolumn{3}{c}{Components} & \multicolumn{2}{l}{IN-A B0 Inc20}  & \multicolumn{2}{l}{VTAB B0 Inc10} \\ 
				\cmidrule(lr){1-3} \cmidrule(lr){4-5}\cmidrule(lr){6-7}
				$L_{bce}$ &$L_{cf}$&$L_{das}$    & {$\bar{\mathcal{A}}$} & ${\mathcal{A}_T}$  & {$\bar{\mathcal{A}}$}  & ${\mathcal{A}_T}$  \\
				\midrule
				\ding{51} & \ding{55} & \ding{55}     & 63.04	          & 53.13         & 94.25	       & 93.28      \\   
				\ding{55} & \ding{51} & \ding{55}     & 67.34      & 57.25      & 95.02    & 94.21   \\ 
				\ding{51} & \ding{51} & \ding{55}     & 67.30       & 57.13     & 94.65    & 93.64      \\ 
				\ding{55} & \ding{55} & \ding{51}     & \bf 68.96  & \bf 59.38  &\bf 95.73  & \bf 95.01      \\ 
				\bottomrule
			\end{tabular}
		}
		
	\end{table}

	\begin{table}[t]
		\caption{\textbf{Generalization study of DAS across five state-of-the-art CIL methods.} The evaluation, performed on ImageNet-A and VTAB using ViT-B/16-IN21K weighs, validates that replacing the conventional cross-entropy loss with DAS yields consistent and significant performance gains.}
		\begin{center}
		\label{tab:ablation-galoss}
		\centering
		\resizebox{1.0\columnwidth}{!}{%
			\begin{tabular}{lcccccc}
				\toprule
				\multicolumn{1}{c}{\multirow{2}{*}{Methods}} 
				& \multicolumn{2}{c}{Components}   & \multicolumn{2}{l}{IN-A B0 Inc20}     & \multicolumn{2}{l}{VTAB B0 Inc10} \\ 
				\cmidrule(lr){2-3} \cmidrule(lr){4-5}\cmidrule(lr){6-7}
				& $L_{ce}$ & $L_{das}$           & {$\bar{\mathcal{A}}$} & ${\mathcal{A}_T}$         & {$\bar{\mathcal{A}}$}  & ${\mathcal{A}_T}$     \\
				\midrule
				ESN~\cite{wang2023isolation}        &\ding{51}  &   \ding{55}      &52.66    & 41.54 & 86.34 & 69.23  \\
				ESN~\cite{wang2023isolation}        &\ding{55}  &   \ding{51}      & \bf53.94    & \bf42.92 & \bf88.77 & \bf72.61  \\    
				\midrule
				RanPAC~\cite{mcdonnell2023ranpac}      &\ding{51}  &   \ding{55}      &70.35     & 62.17  & 92.30   & 91.92  \\
				RanPAC~\cite{mcdonnell2023ranpac}       &\ding{55}  &   \ding{51}      & \bf71.01    & \bf63.06 & \bf93.52  & \bf92.51  \\
				\midrule
				APER~\cite{aper}              &\ding{51}  &   \ding{55}      & 64.63    & 53.85 & 90.20 & 86.16  \\
				APER~\cite{aper}              &\ding{55}  &   \ding{51}      & \bf65.54    & \bf54.25 & \bf92.57 & \bf88.84  \\
				
				\midrule
				EASE~\cite{ease}      &\ding{51}  &   \ding{55}      &65.34    & 55.04 & 93.61 & 93.55  \\
				EASE~\cite{ease}      &\ding{55}  &   \ding{51}      & \bf67.71    & \bf58.85 & \bf95.36  & \bf94.57  \\
				\midrule
				SAFE~\cite{zhao2024safe}      &\ding{51}  &   \ding{55}      &73.14    & 65.90 & 95.22 & 94.91  \\
				SAFE~\cite{zhao2024safe}      &\ding{55}  &   \ding{51}      & \bf73.95    & \bf66.84 & \bf95.96  & \bf95.43  \\
				\bottomrule
			\end{tabular}
		}
		\end{center}
	\end{table}

	\subsection{Comparisons}
	This section presents a comprehensive comparison of DRL with other state-of-the-art CIL methods using ViT-B/16-IN21K or ViT-B/16-IN1K weights on benchmark datasets. 
	
	Table~\ref{tab:benchmark} presents a comparison of DRL against state-of-the-art methods using ViT-B/16-IN21K across six benchmark datasets. DRL consistently outperforms all competing methods, demonstrating superior performance on both standard CIL benchmarks and challenging out-of-distribution datasets. Notably, DRL achieves significant improvements over current state-of-the-art methods including EASE, APER, and DualPrompt across all settings.
	
	On out-of-distribution datasets where domain shift presents substantial challenges, DRL shows approximately 2-4\% improvement over the strongest baseline (EASE). Specifically, on ImageNet-A, VTAB, and ObjectNet, DRL achieves $\bar{\mathcal{A}}$ scores of 68.96\%, 95.73\%, and 72.69\%, outperforming EASE by 3.62\%, 2.12\%, and 1.85\% respectively. These results highlight DRL's exceptional capability to handle distribution shifts while maintaining incremental learning performance.

\begin{figure*}[t]
		\centering
		\includegraphics[width=1.04\textwidth]{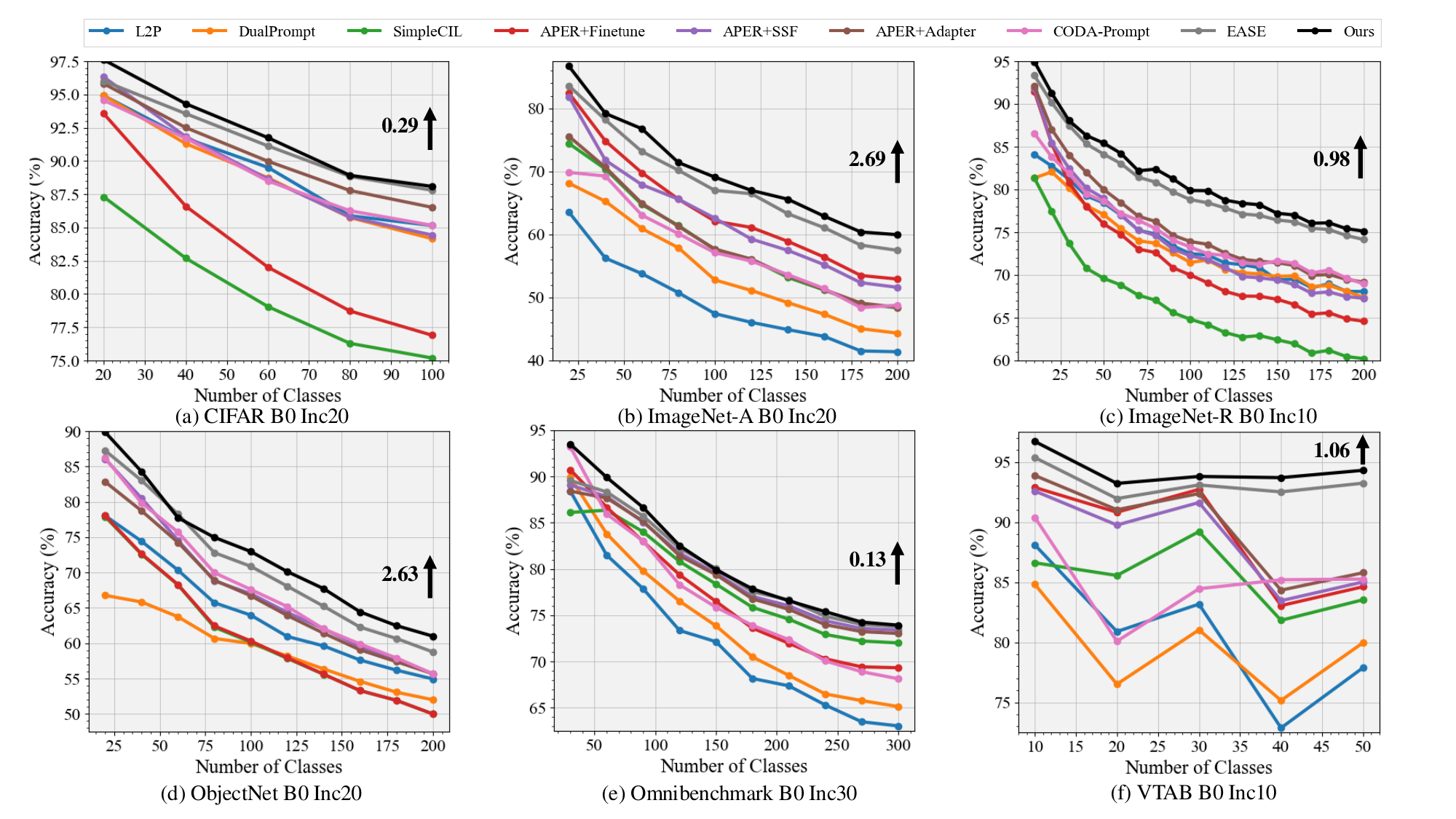}
		\caption{\textbf{Incremental learning performance progression across multiple datasets using ViT-B/16-IN1K weights.} The percentage values indicate DRL's relative improvement over the second-best method at the final task, demonstrating consistent superiority throughout the learning trajectory.}
		
		\label{fig:apple-exps}
	\end{figure*}

	To further validate the generalization capability across different architectures and training configurations,
	we also conduct experiments following APER and EASE using ViT-B/16-IN1K as the pre-trained model under various data splits.
	As shown in Table~\ref{tab:benchmark-typicalmethods} and Fig.~\ref{fig:apple-exps}, 
	DRL maintains competitive performance across all settings. 
	It notably outperforms the second-best method by 2.37\% on ObjectNet and 3.68\% on ImageNet-A, 
	while also showing improvements of 2.69\%, 1.06\%, and 2.63\% on ImageNet-A, VTAB, and ObjectNet respectively compared to EASE. 
	These results consistently demonstrate the robustness and generalizability of our method under diverse experimental conditions.

	\subsection{Ablation Study}
	In this section, we conduct systematic ablation studies to investigate the contribution of each component in DRL, organized around four key aspects: the Decoupled Anchor Supervision mechanism, architectural design choices, hyperparameter sensitivity, and computational efficiency.

	\noindent\textbf{Effect of Decoupled Anchor Supervision (DAS).} 
	We first evaluate the effectiveness of our proposed DAS module in Table~\ref{tab:ablation-loss}. Here, EASE serves as our baseline, and $L_{ce}$ denotes the standard Cross-Entropy loss. Since the total loss of DRL incorporates the knowledge distillation loss $L_{kd}$, we conducted an additional experiment with our baseline to ensure a fair comparison. The results clearly indicate that the proposed DAS is highly effective. In particular, the configuration using DRL without $L_{ce}$ but with $L_{kd}$ and $L_{das}$ achieves the best performance, with $\bar{\mathcal{A}}$ scores of 68.96\% on ImageNet-A, outperforming the setting with only $L_{ce}$ and $L_{kd}$ by 1.72\%. This demonstrates that DAS not only complements but can also effectively replace the conventional cross-entropy loss, leading to robust feature learning in incremental settings.
	
	We further compare DAS against other popular loss functions to highlight its distinctive advantages. As shown in Table~\ref{tab:ablation-loss-others}, we consider Binary Cross-Entropy ($L_{bce}$) and CosFace loss ($L_{cf}$) as competitive alternatives. While combinations of these losses yield reasonable results, DAS alone achieves superior performance across both ImageNet-A and VTAB benchmarks. Specifically, on ImageNet-A, DAS surpasses the best competing loss combination by 1.62\% in terms of $\bar{\mathcal{A}}$, suggesting that its decoupled design provides a more suitable optimization objective for handling the stability-plasticity trade-off in incremental learning.
	
	To validate the general applicability of DAS, we integrate it into five state-of-the-art incremental learning methods: APER, ESN, EASE, RanPAC, and SAFE. The consistent improvements observed in Table~\ref{tab:ablation-galoss} demonstrate DAS's versatility. Across all methods, simply replacing the conventional cross-entropy loss with DAS yields significant performance gains, with improvements ranging from 0.66\% to 2.37\% in $\bar{\mathcal{A}}$ on ImageNet-A. This plug-and-play compatibility, coupled with the absence of additional computational overhead, positions DAS as a valuable component for enhancing existing continual learning frameworks.

	\begin{table}
		\caption{\textbf{Architectural comparison of various attention strategies within the lightweight adapter component of IPA network.} The evaluation includes no attention (\texttt{n-att}), standard self-attention (\texttt{s-att}), and reusable attention (\texttt{r-att}), demonstrating the effectiveness of reusing pre-trained attention patterns for efficient knowledge transfer.}

		\label{tab:ablation-attention}
		\centering
		\resizebox{0.875\columnwidth}{!}{%
			\begin{tabular}{lcccccc}
				\toprule
				\multicolumn{1}{c}{\multirow{2}{*}{Methods}} & \multicolumn{2}{l}{ImageNet-A B0 Inc20} & \multicolumn{2}{l}{VTAB B0 Inc10} \\ 
				\cmidrule(lr){2-3} \cmidrule(lr){4-5}
				& {$\bar{\mathcal{A}}$}             & ${\mathcal{A}_T}$             & {$\bar{\mathcal{A}}$}          & ${\mathcal{A}_T}$          \\
				\midrule
				\texttt{n-att}   & 67.90 & 57.59 & 93.96 & 93.36 \\
				\texttt{s-att}   & 68.78 & \bf59.76 & 95.57 & \bf95.14 \\
				\texttt{r-att}     & \bf68.96 & 59.38 & \bf95.73 & 95.01 \\
				\bottomrule
			\end{tabular}
		}

	\end{table}

	\begin{table}[t]
		\caption{\textbf{Architectural comparison of different transfer gate designs in IPA network.} The evaluation includes direct summation (\texttt{sum}), partial gating (\texttt{gate/part}), adaptive gating (\texttt{gate/adapt}), and extended fusion (\texttt{gate/extra}), demonstrating the optimal balance of performance and efficiency achieved by our adaptive mechanism.}
		\label{tab:gate-analysis}
		\centering
		\resizebox{0.975\columnwidth}{!}{%
			\begin{tabular}{lcccccc}
				\toprule
				\multicolumn{1}{c}{\multirow{2}{*}{Methods}}  & \multicolumn{2}{c}{ImageNet-A B0 Inc20} & \multicolumn{2}{c}{VTAB B0 Inc10} \\ 
				\cmidrule(lr){2-3} \cmidrule(lr){4-5}
				& $\bar{\mathcal{A}}$ & ${\mathcal{A}_T}$ & $\bar{\mathcal{A}}$ & ${\mathcal{A}_T}$ \\
				\midrule
				\texttt{sum} & 66.53 & 56.84 & 94.41 & 93.64 \\               
				\texttt{gate/part} & 67.56 & 57.61 & 94.94 & 93.81 \\
				\texttt{gate/adapt} &  68.96 &  59.38 &  95.73 &  95.01 \\
				\texttt{gate/extra} & 69.21 & 59.62 & 95.80 & 95.13 \\
				\bottomrule
			\end{tabular}%
		}
	\end{table}
	

	\noindent\textbf{Effect of Reusable Attention in IPA.}  
	We then examine the importance of attention mechanisms within the lightweight adapter component of IPA. Table~\ref{tab:ablation-attention} compares three configurations: \texttt{n-att} (no attention modules), \texttt{s-att} (standard self-attention), and our proposed \texttt{r-att} (reusable attention). The empirical results demonstrate that incorporating attention substantially enhances model performance, with both \texttt{s-att} and \texttt{r-att} outperforming the attention-free baseline by approximately 1-1.5\% in $\bar{\mathcal{A}}$ on ImageNet-A.
	The key innovation of our reusable attention mechanism lies in its ability to leverage the pre-trained model's original attention matrix $\mathbf{A}^o$, which inherently captures rich feature interdependencies learned during large-scale pre-training. This approach provides two significant advantages: first, it preserves the model's {plasticity} by maintaining the semantic relationships encoded in the pre-trained attention patterns; second, it eliminates additional parametric overhead since no new attention parameters need to be learned. The performance gap between \texttt{r-att} (68.96\%) and \texttt{s-att} (68.78\%) on ImageNet-A, though modest, confirms that distilled attention knowledge can be effectively transferred to downstream adaptation modules while maintaining efficiency. This architectural innovation demonstrates that careful reuse of pre-existing components can enhance functionality without compromising computational efficiency.
	

	\noindent\textbf{Effect of Transfer Gate in IPA.} 
	We further investigate different feature fusion strategies for the transfer gate, which plays a critical role in balancing current and historical information flow. 
	Table~\ref{tab:gate-analysis} compares four approaches: direct summation (\texttt{sum}) defined as $\boldsymbol{f}_t^{e_l} = \boldsymbol{f}_t^{\bar{e}_l} + \boldsymbol{f}_{t-1}^{g_l}$; partial gating (\texttt{gate/part}) formulated as $\boldsymbol{f}_t^{e_l} = \boldsymbol{f}_t^{\bar{e}_l} + \mathbf{M}_t^l \boldsymbol{f}_{t-1}^{g_l}$; adaptive gating (\texttt{gate/adapt}) expressed as $\boldsymbol{f}_t^{e_l} = (1-\mathbf{M}_t^l)\boldsymbol{f}_t^{\bar{e}_l} + \mathbf{M}_t^l \boldsymbol{f}_{t-1}^{g_l}$; and extended fusion (\texttt{gate/extra}) computed as $\boldsymbol{f}_t^{e_l} = \frac{1}{2}\left[(2-\mathbf{M}_t^l - \mathbf{M}_0^l)\boldsymbol{f}_t^{\bar{e}_l} + \mathbf{M}_t^l\boldsymbol{f}_{t-1}^{e_l} + \mathbf{M}_0^l\boldsymbol{f}_{0}^{o_l}\right]$. 
	Our systematic exploration establishes the adaptive gating mechanism (\texttt{gate/adapt}) as the optimal solution, delivering superior performance without parameter overhead. As Table~\ref{tab:gate-analysis} demonstrates, it achieves significant improvements over alternatives: +2.43\% $\bar{\mathcal{A}}$ versus direct summation (\texttt{sum}) and +1.40\% over partial gating (\texttt{gate/part}) on ImageNet-A, while maintaining full parameter efficiency. 
	The extended fusion variant (\texttt{gate/extra}) yields only marginal gains (0.25\% $\bar{\mathcal{A}}$ improvement) at substantial cost—introducing 0.29\% additional parameters and increased architectural complexity. Crucially, our adaptive gating solution provides three decisive advantages: superior parameter efficiency requiring zero additional parameters beyond the base architecture; optimal feature integration balancing historical and current knowledge; and enhanced stability-plasticity tradeoff through dynamic feature weighting. 
	The 2.43\% performance leap over direct summation validates our core design principle: selective feature integration through learnable masks is fundamental for effective incremental learning. While future work may explore hierarchical formulations like $\boldsymbol{f}_t^{e_l} = \frac{1}{t+1}[(t+1-\sum_{i=1}^{t-1}\mathbf{M}_i^l - \mathbf{M}_0^l)\boldsymbol{f}_t^{\bar{e}_l} + \sum_{i=1}^{t-1}\mathbf{M}_i^l\boldsymbol{f}_{i}^{e_l}+\mathbf{M}_0^l\boldsymbol{f}_{0}^{o_l}]$, these inevitably increase complexity. Our adaptive gating remains the preferred solution, achieving the optimal equilibrium between accuracy (95.73\% $\bar{\mathcal{A}}$ on VTAB), efficiency, and architectural elegance.

	
\begin{figure*}[t] 
	\centering
	\subfloat[$T=1$ w/o DAS.\label{fig:our-ce-visual-stage1}]{
		\includegraphics[width=0.23\textwidth]{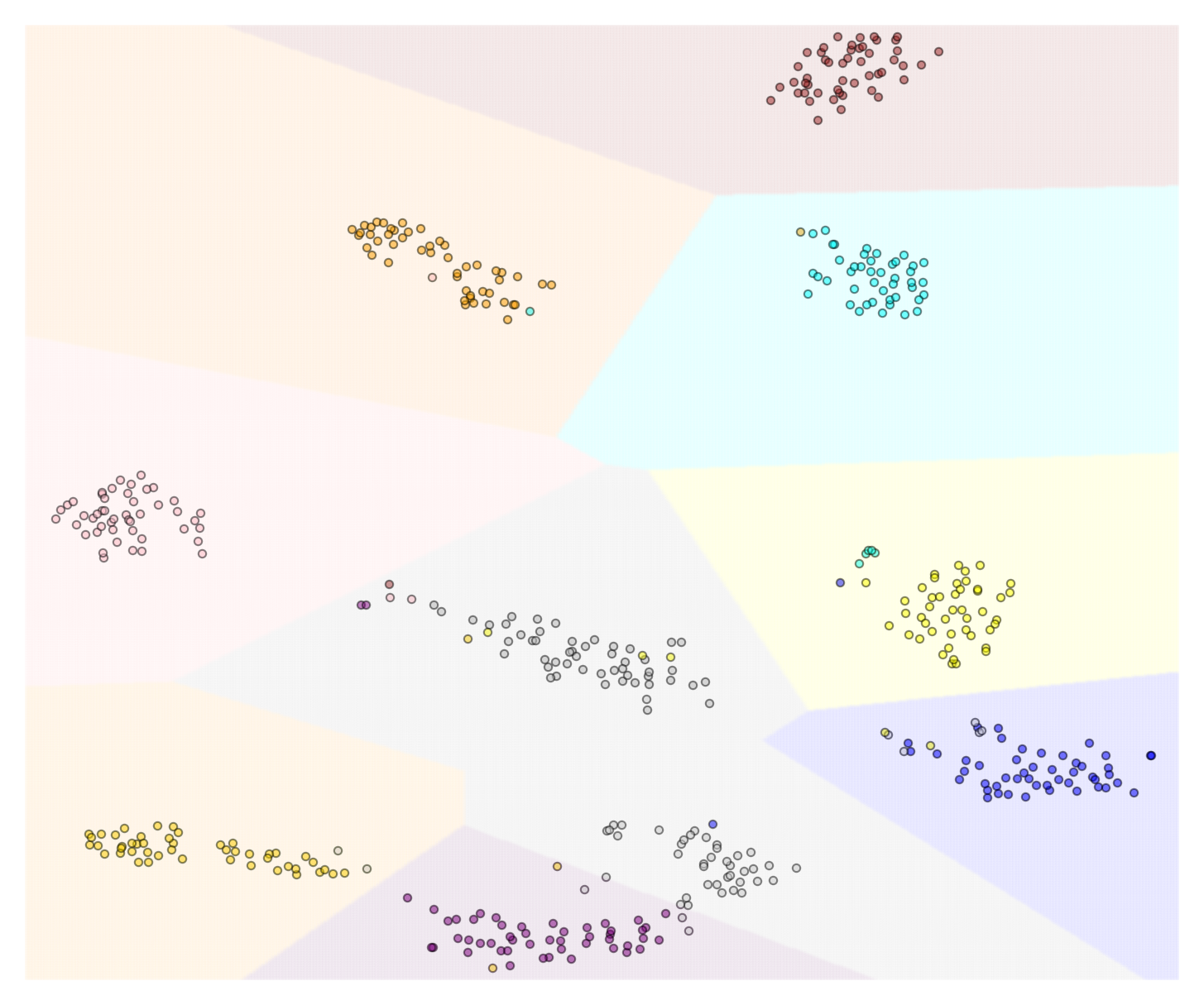}
	}
	\hfill
	\subfloat[$T=2$ w/o DAS.\label{fig:our-ce-visual-stage2}]{
		\includegraphics[width=0.23\textwidth]{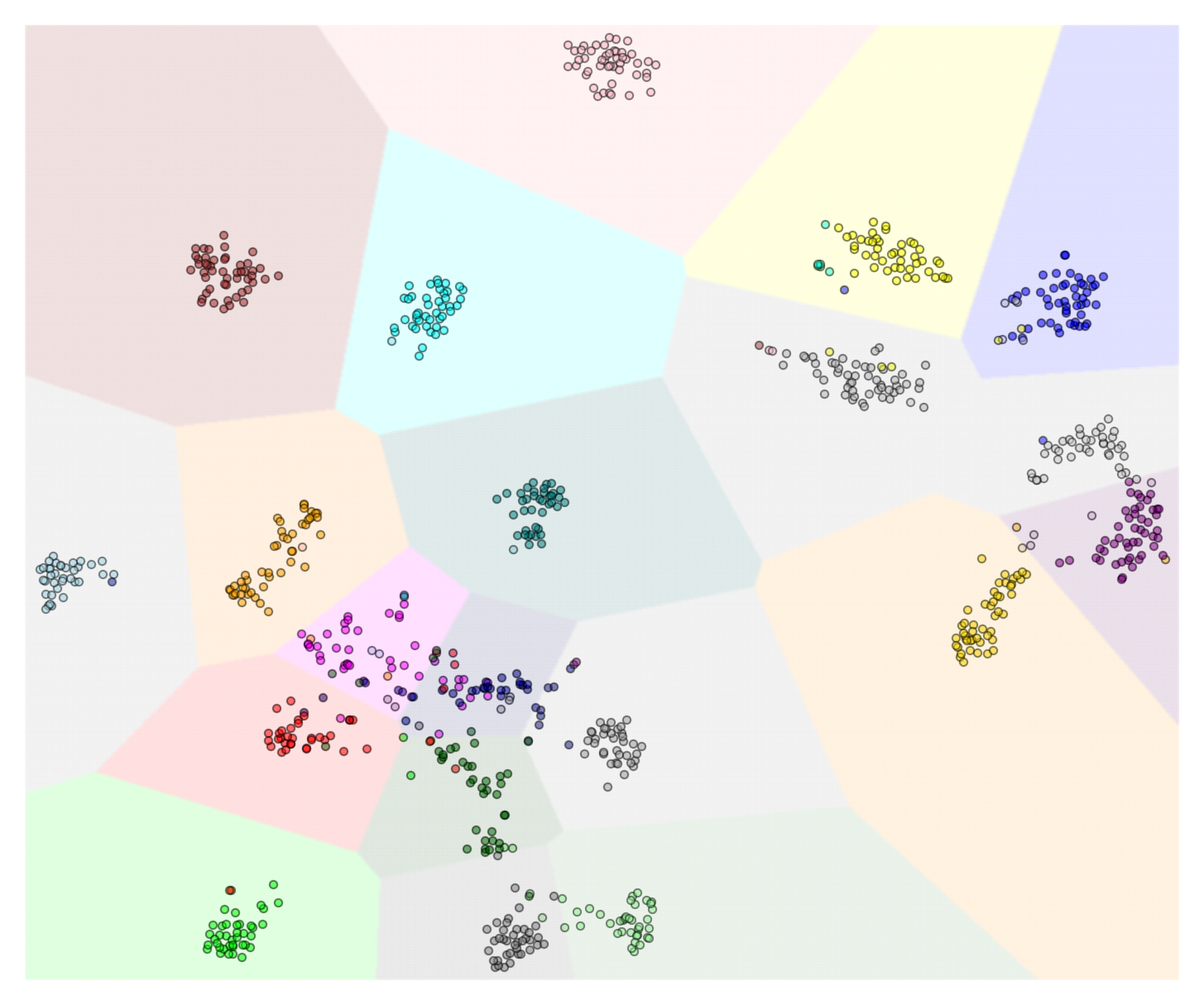}
	}
	\hfill
	\subfloat[$T=1$ with DAS.\label{fig:our-ga-visual-stage1}]{
		\includegraphics[width=0.23\textwidth]{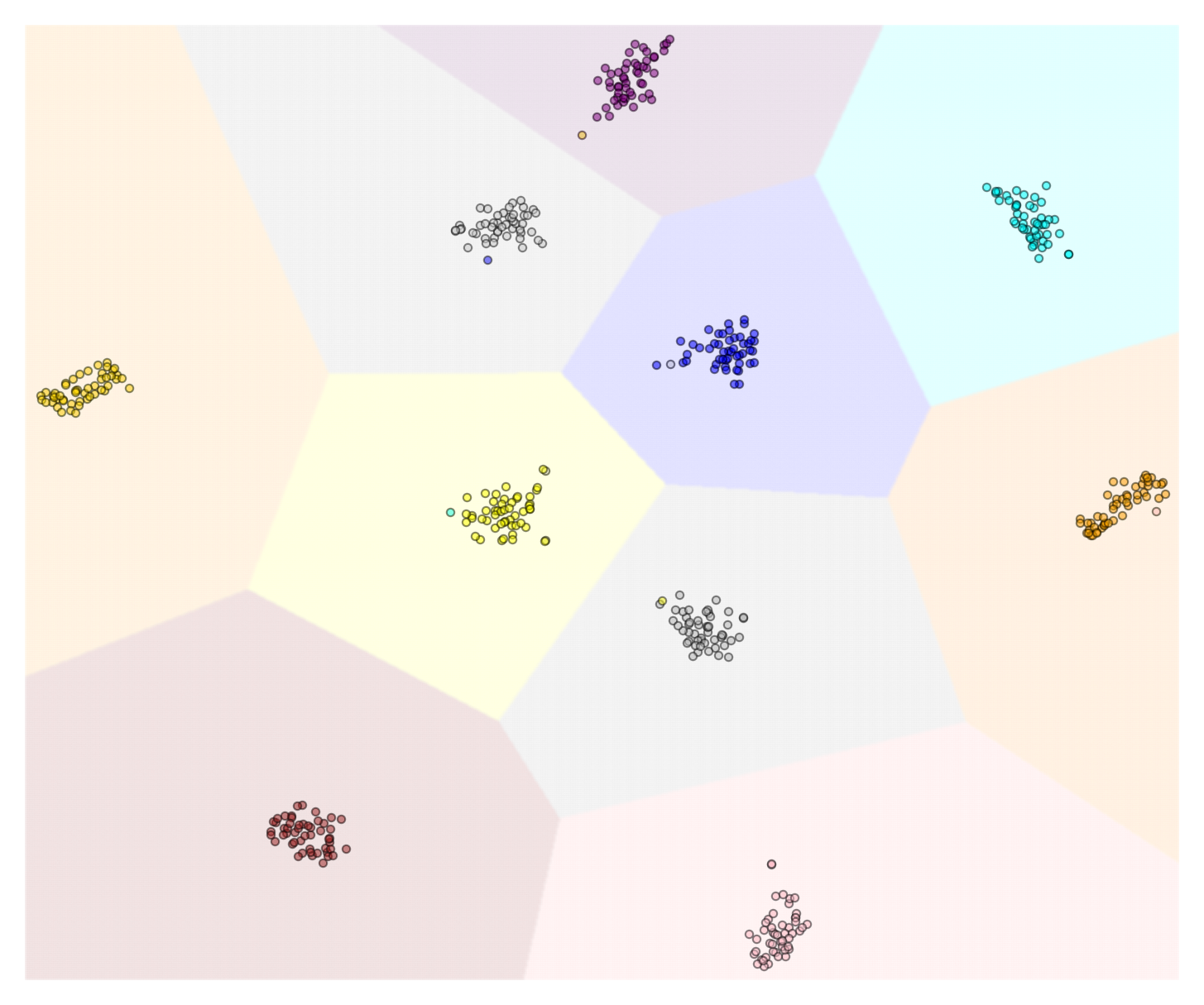}
	}
	\hfill
	\subfloat[$T=2$ with DAS.\label{fig:our-ga-visual-stage2}]{
		\includegraphics[width=0.23\textwidth]{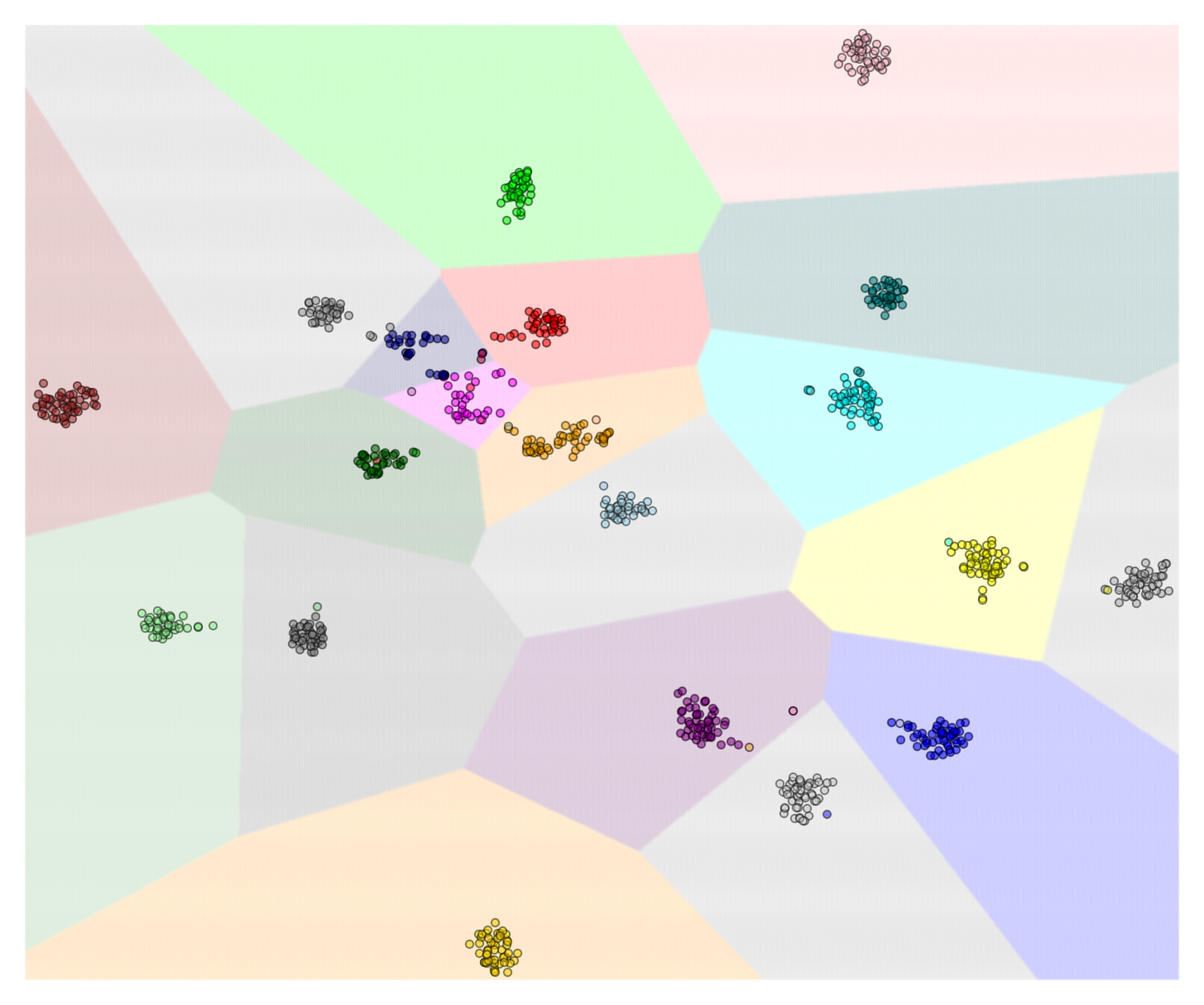}
	}
	\caption{\textbf{t-SNE visualizations of feature representations on VTAB dataset with B0 Inc10 setting,} comparing the effect of Decoupled Anchor Supervision (DAS) on feature separability and cluster compactness across incremental learning stages.}
	\label{fig:main_fig}
\end{figure*}
	
	
	\begin{table}[t]
		\centering
		\caption{\textbf{Architectural analysis of bottleneck widths in the final transformer block of IPA network.} The evaluation compares various bottleneck configurations (\texttt{768→r→768}) with different hidden dimensions \texttt{r}, demonstrating the trade-off between model complexity and representation capacity in adapter-based continual learning.}
		\resizebox{\columnwidth}{!}{%
			\begin{tabular}{lcccccccc}
				\toprule
				\multicolumn{1}{c}{\multirow{2}{*}{Architecture}} & \multicolumn{2}{l}{ImageNet-A B0 Inc20} & \multicolumn{2}{l}{VTAB B0 Inc10} \\ 
				\cmidrule(lr){2-3} \cmidrule(lr){4-5}
				& {$\bar{\mathcal{A}}$}   & ${\mathcal{A}_T}$     & {$\bar{\mathcal{A}}$}     & ${\mathcal{A}_T}$      \\
				\midrule
				\texttt{768→192→768}  & 68.81 & 59.03 & 95.52 & 94.88 \\
				\texttt{768→384→768}  & 68.78 & 59.04 & 95.51 & 94.92 \\
				\texttt{768→768→768} & \textbf{68.96} & 59.38 & \textbf{95.73} & 95.01 \\
				\texttt{768→1536→768} & 68.98 & 59.24 & 95.63 & \textbf{95.14} \\
				\texttt{768→2304→768} & 68.98 & \textbf{59.83} & 95.56 & 94.94 \\
				\texttt{768→3072→768} & 68.97 & 59.82 & 95.61 & 94.99 \\
				\bottomrule
			\end{tabular}
		}
		\label{tab:ablation-lastlayer}
	\end{table}
	
	\noindent\textbf{Architectural Optimization in Final Block.}
	\label{supple:2linear}
	We analyze the design choices for the linear transformations within the IPA network's final block, where we replace the original Feedforward Network (FFN) with two lightweight linear layers. Table~\ref{tab:ablation-lastlayer} compares various dimensionality configurations, where \texttt{768→384→768} indicates a bottleneck architecture with hidden dimension 384, and the original FFN is represented as \texttt{768→3072→768} (standard ViT expansion factor of 4).
	Our experiments reveal that the configuration \texttt{768→768→768} achieves an optimal balance between performance and parameter efficiency. While larger hidden dimensions (\texttt{1536}, \texttt{2304}, \texttt{3072}) show marginal improvements in certain metrics, the gains are insufficient to justify the substantial parameter increase. For instance, expanding to \texttt{768→3072→768} increases parameters by 4× but does not improves $\bar{\mathcal{A}}$ compared to our chosen configuration. This suggests that for adapter-based continual learning, maintaining dimensional consistency with the pre-trained features (\texttt{768} dimensions) provides sufficient representational capacity without introducing unnecessary complexity. This optimization strategy reduces model complexity compared to the standard FFN while maintaining competitive performance, demonstrating that efficient adapter design is crucial for practical continual learning systems.

	\begin{table}[t]
		\caption{\textbf{Sensitivity analysis of loss weighting parameters in our DRL framework.} The study systematically assesses the knowledge distillation weight ($\alpha$), positive constraint weight ($\lambda_p$), and negative constraint weight ($\lambda_n$), revealing robust performance across a wide range of values and identifying the optimal configuration for task performance.}
		
		\label{tab:ablation-lossweight}
		\centering
		\resizebox{1.0\columnwidth}{!}{%
			\begin{tabular}{ccccccc}
				\toprule
				\multicolumn{1}{c}{\multirow{2}{*}{$\alpha$}} &  \multicolumn{1}{c}{\multirow{2}{*}{$\lambda_p$}} &\multicolumn{1}{c}{\multirow{2}{*}{$\lambda_n$}} &
				\multicolumn{2}{c}{ImageNet-A B0 Inc20} & \multicolumn{2}{c}{VTAB B0 Inc20}  \\
				\cmidrule(lr){4-5}\cmidrule(lr){6-7}
				& & & {$\bar{\mathcal{A}}$} & ${\mathcal{A}_T}$  & {$\bar{\mathcal{A}}$} & ${\mathcal{A}_T}$	\\
				\midrule
				0   & 1   & 2   & 67.29 & 57.01 & 94.90 & 93.97 \\
				0.5 & 1   & 2   & \bf 68.96  & \bf 59.38 & \bf 95.73 & \bf 95.01 \\
				1   & 1   & 2   & 68.74 & 58.72 & 95.21 & 94.30  \\
				3   & 1   & 2   & 68.11 & 57.47 & 94.73  & 93.72 \\
				5   & 1   & 2   & 67.51 & 56.35 & 94.17 & 93.22 \\
				\midrule
				0.5 & 1& 0.5 & 65.52  & 54.97 & 94.65 & 93.75 \\
				0.5 & 1  & 1   & 68.46  & 58.07 & 95.00     & 94.16 \\
				0.5 & 1  & 3   & 68.17  & 58.06 & 95.40 & 94.49 \\
				0.5 & 1 & 5   & 67.16  & 57.47 & 94.93 & 94.16 \\		
				\midrule
				0.5 & 0.5& 1 & 66.72  & 56.93 & 94.84 & 94.35 \\
				0.5 & 1  & 1   & 68.46  & 58.07 & 95.00     & 94.16 \\
				0.5 & 3  & 1   & 68.91  & 59.06 & 95.62 & 94.97 \\
				0.5 & 5 & 1   & 67.54  & 56.38 & 94.84 & 94.07 \\
				\bottomrule
			\end{tabular}			
		}
	\end{table}

	\noindent\textbf{Hyperparameter Sensitivity.} 
	We conduct extensive sensitivity analysis on key hyperparameters to assess model robustness. Table~\ref{tab:ablation-anchor} examines the impact of virtual anchor parameter $k$, showing stable performance within the range [0,2]. It is important to clarify that in our current implementation, we employ a single shared anchor $k$ for both positive and negative constraints, as defined by $z_i>k$ for positive samples and $z_j<-k$ for negative samples. This symmetric design provides a balanced separation boundary while maintaining simplicity.

	When $k=0$, the anchor margin becomes zero, providing the most direct form of decoupled supervision where positive and negative constraints are completely separated without additional margin. 
	At $k=1$, the predefined virtual anchor introduces a positive margin that further enhances feature discriminability by creating clearer separation boundaries between classes. This margin acts as a fixed reference point across all incremental stages, enforcing consistent logit interpretation and feature space alignment. The performance improvement from $k=0$ to $k=1$ confirms that the margin provides additional benefits for feature separation beyond the basic decoupling mechanism. While DAS can be extended to use separate anchors $k_+$ and $k_-$ for positive and negative constraints respectively, our experiments focus on the symmetric case for simplicity and stability.
	
	However, excessively large $k$ values ($\geq$3) gradually degrade performance, as overly aggressive margin settings may impede intra-class cohesion by introducing excessive separation requirements. The supervision signal becomes too stringent, making it difficult to maintain compact class distributions while satisfying the margin constraints. This demonstrates the importance of maintaining an appropriate balance between leveraging margin-based separation and preserving learnable class distributions.
	
	Similarly, Table~\ref{tab:ablation-lossweight} analyzes the sensitivity to loss weighting parameters $\alpha$, $\lambda_p$, and $\lambda_n$. The results indicate robust performance across a wide range of values, with optimal performance achieved at $\alpha=0.5$, $\lambda_p=3$, $\lambda_n=1$. The model shows particular stability for $\lambda_p, \lambda_n \in [1, 3]$, with performance variations of less than 0.5\% within this range. This robustness is advantageous for practical applications where extensive hyperparameter tuning may be impractical. The sensitivity analysis collectively demonstrates that DRL provides consistent performance across reasonable parameter choices, reducing the burden of meticulous hyperparameter optimization while maintaining competitive performance through its well-designed architectural components.

	\subsection{Extended Analysis}
	\noindent\textbf{Parameter and Computational Efficiency.} 
	To illustrate the efficiency of DRL, we provide a comprehensive analysis of parameter counts and computational requirements during both training and inference phases, as summarized in Fig.~\ref{fig:main_fig}. 
	The baseline ViT-B/16 model contains approximately 1B parameters (denoted as `1B'). Our DRL approach introduces only 0.6\% additional trainable parameters relative to the base model, demonstrating exceptional parameter efficiency. During inference, the total parameter footprint scales minimally to (1 + 0.006$t$)B, where $t$ represents the number of training stages, ensuring efficiency in both training and deployment phases.
	In terms of specific parameter counts, the pre-trained model (PTM) contains $\sim$102M parameters. For comparison, EASE introduces $\sim$0.29M learnable parameters, while our IPA network requires $\sim$0.62M learnable parameters. Notably, both methods maintain learnable parameters below 1\% of the base model size, ensuring lightweight adaptation with minimal storage overhead.
	A key computational advantage of our approach lies in the inference process. While EASE requires executing inference through all previously trained models ($T$ times) to obtain the final feature representation $\mathbf{F}_{T}$, our DRL approach performs inference only once through the final integrated model. This single-pass inference strategy substantially reduces computational overhead, memory requirements, and latency compared to methods requiring multiple forward passes. The combination of minimal parameter growth and efficient single-pass inference makes DRL particularly suitable for resource-constrained deployment scenarios where both storage and computational resources are limited.

	\begin{table}[t]
		\caption{\textbf{Sensitivity analysis of virtual anchor parameter $k$ in Decoupled Anchor Supervision.} The evaluation  reveals performance stability across a wide range of anchor values, highlighting the importance of appropriate margin settings for effective feature separation and cross-stage alignment.}
		
		\label{tab:ablation-anchor}
		\centering
		\resizebox{0.95\columnwidth}{!}{%
			\begin{tabular}{ccccccc}
				\toprule
				\multicolumn{1}{c}{\multirow{2}{*}{$k$}} & \multicolumn{3}{c}{IN-A B0 Inc20}  & \multicolumn{3}{c}{VTAB B0 Inc10} \\ 
				\cmidrule(lr){2-4} \cmidrule(lr){5-7}
				& s & {$\bar{\mathcal{A}}$}             & ${\mathcal{A}_T}$           & s & {$\bar{\mathcal{A}}$}          & ${\mathcal{A}_T}$          \\
				\midrule
				0.0    & 12.47  & 68.67 & 58.53 & 10.11 & 94.96 & 94.08 \\  
				0.5  & 13.75  & 68.81 & 58.72 & 11.29 & 95.21 & 94.47 \\  
				1.0    & 14.95  & \bf68.96 & \bf59.38 & 12.46 & \bf95.73 & \bf95.01 \\  
				2.0    & 17.33  & 68.58 & 59.18 & 14.77 & 94.99 & 94.23 \\  
				3.0   & 19.53  & 67.85 & 57.60 & 17.00 & 94.35 & 93.51 \\  
				5.0    & 23.60  & 65.16 & 55.83 & 21.49 & 94.05 & 93.04 \\  
				\bottomrule
			\end{tabular}
		}
	\end{table}

\begin{figure}[t]
	\centering
	\begin{minipage}[b]{0.23\textwidth}
		\centering
		\subfloat[\centering{Stage 1 feature embeddings\newline on CIFAR100}\label{fig:our-ga-cifar-stage1}]{
			\includegraphics[width=0.975\textwidth]{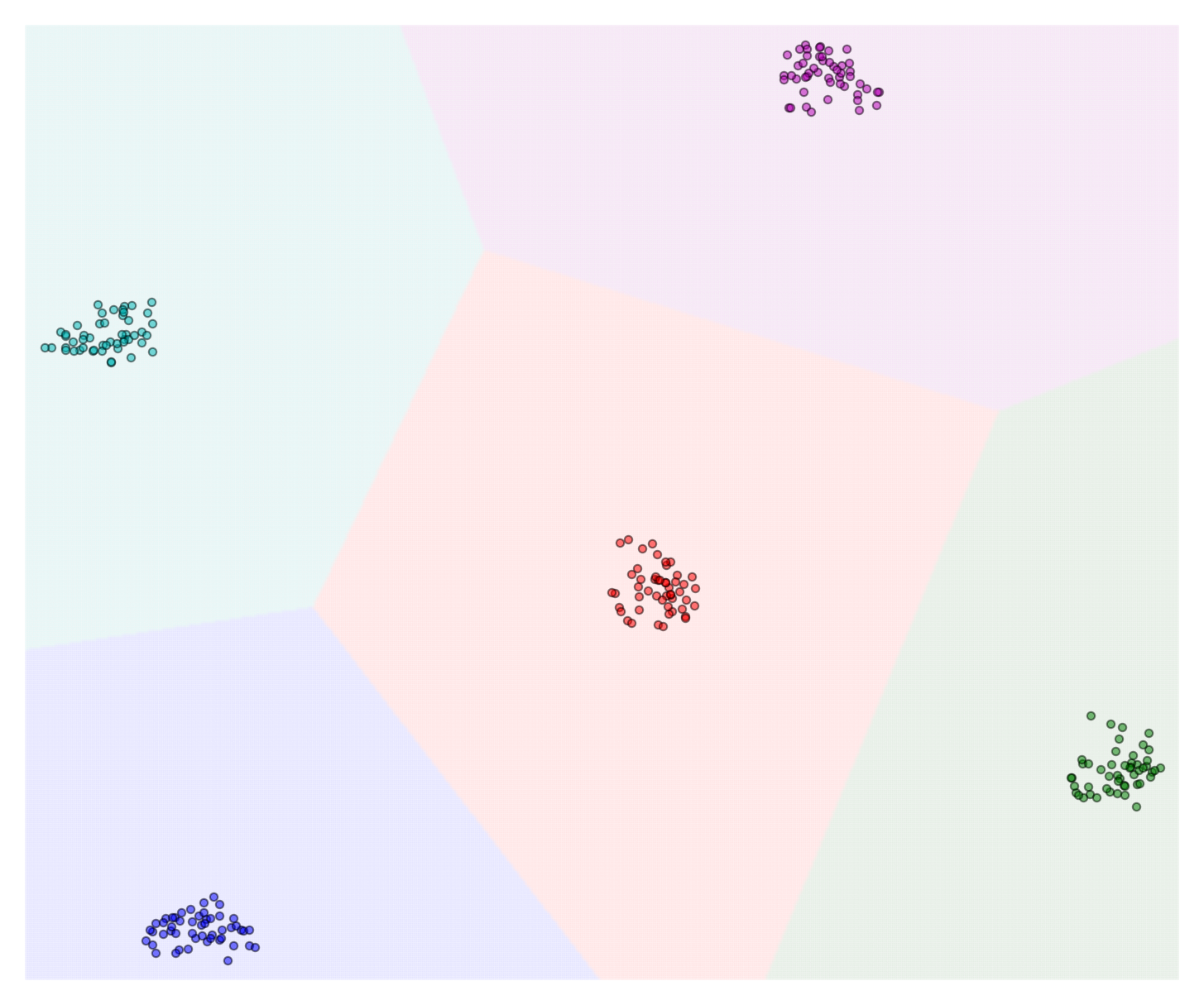}
		}
	\end{minipage}
	\hfill
	\begin{minipage}[b]{0.23\textwidth}
		\centering
		\subfloat[\centering{Stage 2 feature embeddings\newline on CIFAR100}\label{fig:our-ga-cifar-stage2}]{
			\includegraphics[width=0.975\textwidth]{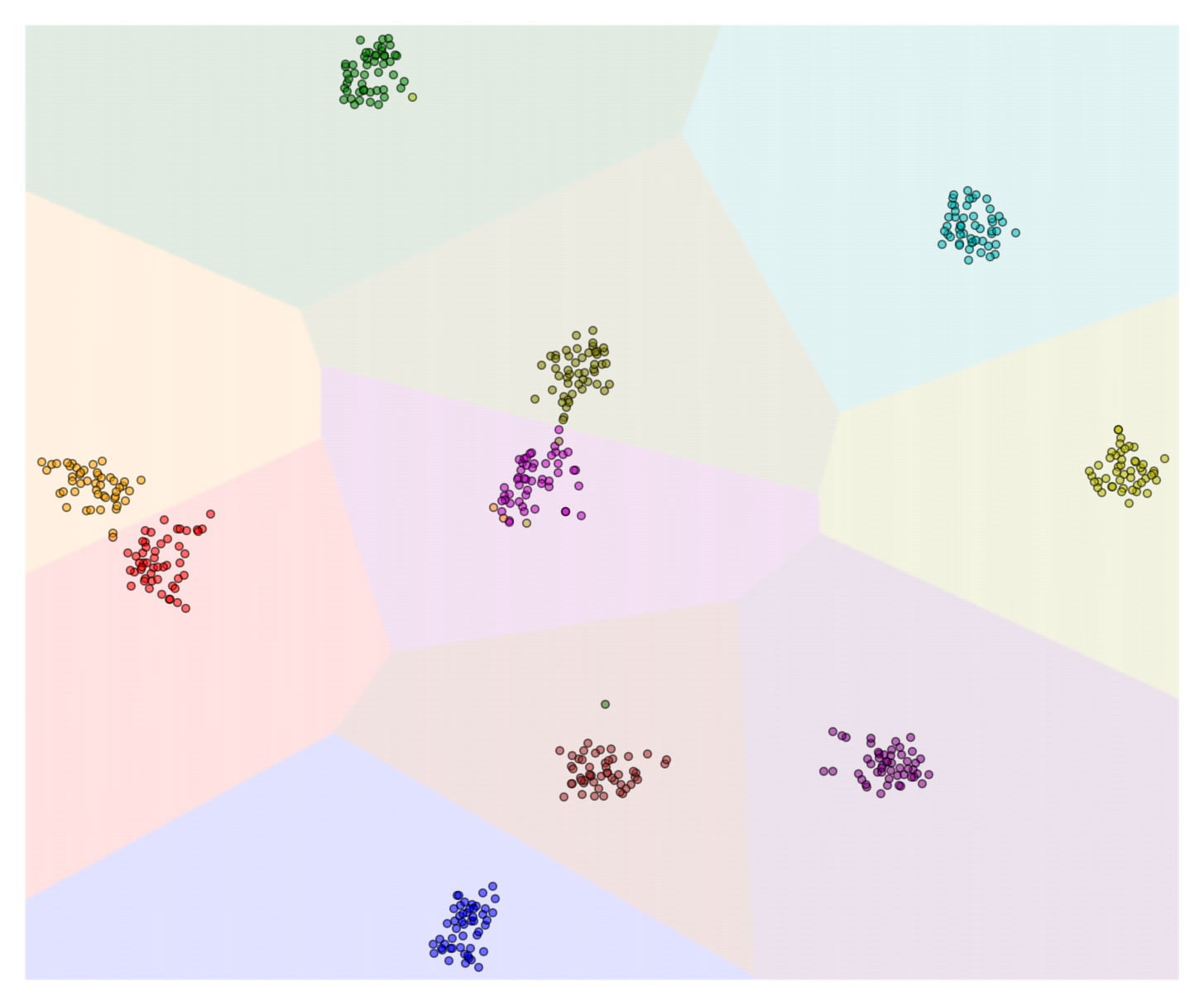}
		}
	\end{minipage}
	\caption{\textbf{t-SNE visualizations of DRL embeddings on CIFAR100 under `B0 Inc5' incremental setting.}}
	\label{fig:main_fig2}
\end{figure}

	
	\noindent\textbf{Visualization.}
	To qualitatively assess representation quality, we leverage t-SNE~\cite{van2008visualizing} to visualize the learned decision boundaries on the VTAB dataset across incremental stages. Fig.~\ref{fig:main_fig} illustrates the feature distributions on the VTAB dataset under the B0 Inc10 setting, comparing configurations with and without DAS. The visualizations reveal that DRL with DAS produces more compact and well-separated class clusters compared to the variant using conventional cross-entropy loss. Specifically, at stage 2 (Fig.~\ref{fig:our-ga-visual-stage2}), the DAS-enhanced features maintain clear separation boundaries between classes from both incremental stages, demonstrating effective mitigation of catastrophic forgetting while accommodating new concepts.
	Additional visualizations on CIFAR100 (Fig.~\ref{fig:main_fig2}) under the B0 Inc5 setting further confirm these observations. The t-SNE plots show that DRL effectively distinguishes instances into their respective classes with minimal inter-class overlap. The superior clustering quality visually corroborates the quantitative performance advantages observed in our benchmark evaluations, providing intuitive evidence that DAS promotes more discriminative feature learning after the incremental learning process. These visualizations collectively demonstrate that our method successfully addresses the fundamental challenge of maintaining representation quality while sequentially incorporating new knowledge.

	\section{Limitations and Future Work}
	While DRL demonstrates strong performance across multiple benchmarks, several limitations present opportunities for future research. The current transfer gate design, though effective, could benefit from more sophisticated feature fusion mechanisms such as attention-based dynamic weighting or hierarchical gating structures that adapt to class complexity. The virtual anchor parameter $k$ is currently fixed throughout training; developing adaptive sampling strategies that adjust $k$ based on class complexity or data distribution could further enhance performance. Additionally, while our experiments focus on medium-scale models, extending validation to larger architectures (e.g., ViT-L, ViT-H) would strengthen the evidence for scalability. Future work will also explore biological plausibility of the distillation mechanism and investigate applications in more challenging continual learning scenarios including task-incremental and domain-incremental settings. We will release code to facilitate reproduction and adoption by the research community.

	\section{Conclusion}
	In this paper, we proposed a novel class-incremental learning (CIL) method, Discriminative Representation Learning (DRL), which consists of an Incremental Parallel Adapter (IPA) network and Decoupled Anchor Supervision (DAS). The IPA network achieves high efficiency and smooth representation shift through a lightweight adapter and a learnable transfer gate, enabling gradual model expansion while maintaining parameter efficiency. DAS enhances discriminative representation learning by decoupling constraints for positive and negative samples via a fixed anchor, which helps align feature spaces across incremental stages and mitigates optimization-inference inconsistency. Extensive experiments on six benchmarks demonstrate that DRL achieves state-of-the-art performance while maintaining high parameter efficiency.
	
	\balance
	
	\bibliographystyle{IEEEtran}
	\bibliography{bare_jrnl_new_sample4}

\end{document}